\documentclass{article}

\usepackage{microtype}
\usepackage{graphicx}
\usepackage{subcaption}
\usepackage{booktabs} 

\usepackage{hyperref}

\usepackage[preprint]{icml2026}

\usepackage{amsmath}
\usepackage{amssymb}
\usepackage{mathtools}
\usepackage{amsthm}
\usepackage{CJKutf8}

\usepackage[capitalize,noabbrev]{cleveref}

\usepackage{multirow}
\usepackage{multicol}
\usepackage{booktabs}
\usepackage{graphicx}
\usepackage{tikz}
\usepackage{paralist}
\usepackage{booktabs}
\usepackage{amsfonts}
\usepackage{CJKutf8}
\usepackage{amssymb}
\usepackage{enumitem}
\usepackage{setspace}
\usepackage{amsmath} 
\usepackage{amssymb}
\usepackage{xcolor}
\usepackage{graphicx}
\usepackage{url}
\usepackage{booktabs}
\usepackage{amssymb}
\usepackage{bbding}
\usepackage{pifont}
\usepackage{wasysym}
\usepackage{utfsym}
\usepackage{fontawesome}
\usepackage{footnote} 
\usepackage{makecell}
\usepackage{wrapfig}
\usepackage{ulem}

\usepackage{CJKutf8}

\theoremstyle{plain}

\theoremstyle{definition}

\theoremstyle{remark}

\usepackage[textsize=tiny]{todonotes}

\icmltitlerunning{Tracing Meta-Cognitive Activation Trajectory in R1-Style LLMs}

\begin{document}

\twocolumn[
  \icmltitle{From Latent Signals to Reflection Behavior:\\ Tracing Meta-Cognitive Activation Trajectory in R1-Style LLMs}



  \icmlsetsymbol{equal}{*}
  \icmlsetsymbol{cor}{$\dagger$}

  \begin{icmlauthorlist}
    \icmlauthor{Yanrui Du}{equal,scir}
    \icmlauthor{Yibo Gao}{equal,scir}
    \icmlauthor{Sendong Zhao}{cor,scir}
    \icmlauthor{Jiayun Li}{scir}
    \icmlauthor{Haochun Wang}{scir}
    \\
    \icmlauthor{Qika Lin}{nus}
    \icmlauthor{Kai He}{nus}
    \icmlauthor{Bing Qin}{scir}
    \icmlauthor{Mengling Feng}{nus}
  \end{icmlauthorlist}

  \icmlaffiliation{scir}{SCIR Lab, Harbin Institute of Technology}
  \icmlaffiliation{nus}{National University of Singapore}

  \icmlcorrespondingauthor{Sendong Zhao}{sdzhao@ir.hit.edu.cn}

  \icmlkeywords{Machine Learning, ICML}

  \vskip 0.3in
]

\printAffiliationsAndNotice{%
  \textsuperscript{*}Equal contribution.
  \textsuperscript{$\dagger$}Corresponding author.
  Emails: \{yrdu,sdzhao\}@ir.hit.edu.cn.
}




\begin{abstract}
R1-style LLMs have attracted growing attention for their capacity for self-reflection, yet the internal mechanisms underlying such behavior remain unclear. 
To bridge this gap, we anchor on the onset of reflection behavior and trace its layer-wise activation trajectory. 
Using the logit lens to read out token-level semantics, we uncover a structured progression: (i) Latent-control layers, where an approximate linear direction encodes the semantics of thinking budget; 
(ii) Semantic-pivot layers, where discourse-level cues, including turning-point and summarization cues, surface and dominate the probability mass; 
and (iii) Behavior-overt layers, where the likelihood of reflection-behavior tokens begins to rise until they become highly likely to be sampled.
Moreover, our targeted interventions uncover a causal chain across these stages: prompt-level semantics modulate the projection of activations along latent-control directions, thereby inducing competition between turning-point and summarization cues in semantic-pivot layers, which in turn regulates the sampling likelihood of reflection-behavior tokens in behavior-overt layers.
Collectively, our findings suggest a human-like meta-cognitive process—progressing from latent monitoring, to discourse-level regulation, and to finally overt self-reflection.
Our analysis code can be found at \url{https://github.com/DYR1/S3-CoT}.
\end{abstract}

\vspace{-0.9cm}
\section{Introduction}

R1-style Large Language Models (LLMs) have drawn growing attention for their apparent self-reflection behaviors, such as emitting markers like ``Wait'' and ``Hmm''~\cite{yang2025qwen3,jaech2024openai,guo2025deepseek}. 
Yet the internal mechanisms behind such human-like behaviors remain poorly understood. 
This gap matters beyond interpretability for its own sake: without a mechanistic view, reflection behavior is hard to predict, control, or reliably improve. 
Accordingly, our study offers a principled framework for linking internal representations to discourse-level decisions and, ultimately, to user-visible self-correction.

Early efforts to elicit reflection behavior have largely relied on human meta-cognitive intuitions to design heuristic frameworks~\cite{wei2022chain,shinn2023reflexion}.
These approaches are fundamentally behavior-centric, building agentic and search-based inference procedures that impose external structure for planning and iterative revision (e.g., ReAct, Tree-of-Thoughts, and Reflexion)~\cite{yao2022react,yao2023tree,madaan2023self}. 
With the emergence of DeepSeek-R1, reflection has been internalized as a default capacity—shifting from inference-time scaffolding toward mechanisms embedded within the model itself~\cite{wang2025wait,arora2025training,chen2025learning}. 
This shift raises a central question: as a higher-order cognitive capacity in humans~\cite{fleming2012neural,dunlosky2008metacognition}, how does reflection unfold within an LLM's internal mechanisms?
While recent studies suggest that thinking-related signals emerge in LLM representations~\cite{zhu2025emergence,cabannes2024iteration,zhang2025reasoning,chen2024language}, they often stop short of establishing a stage-wise, causal explanation—pinpointing which layers are uniquely responsible, how signals propagate across depth, and how they ultimately give rise to explicit reflection markers. 
By contrast, mechanistic analyses of other behavioral attributes (e.g., safety and sentiment) have begun to map clearer causal chains within forward computation~\cite{zhou2024alignment,tak2025mechanistic,zou2023representation}. 
These advances motivate us to pursue a comprehensive, mechanism-oriented study of reflective behavior in R1-style LLMs.

\begin{figure*}[ht]
  \centering
    \includegraphics[width=0.88\linewidth]{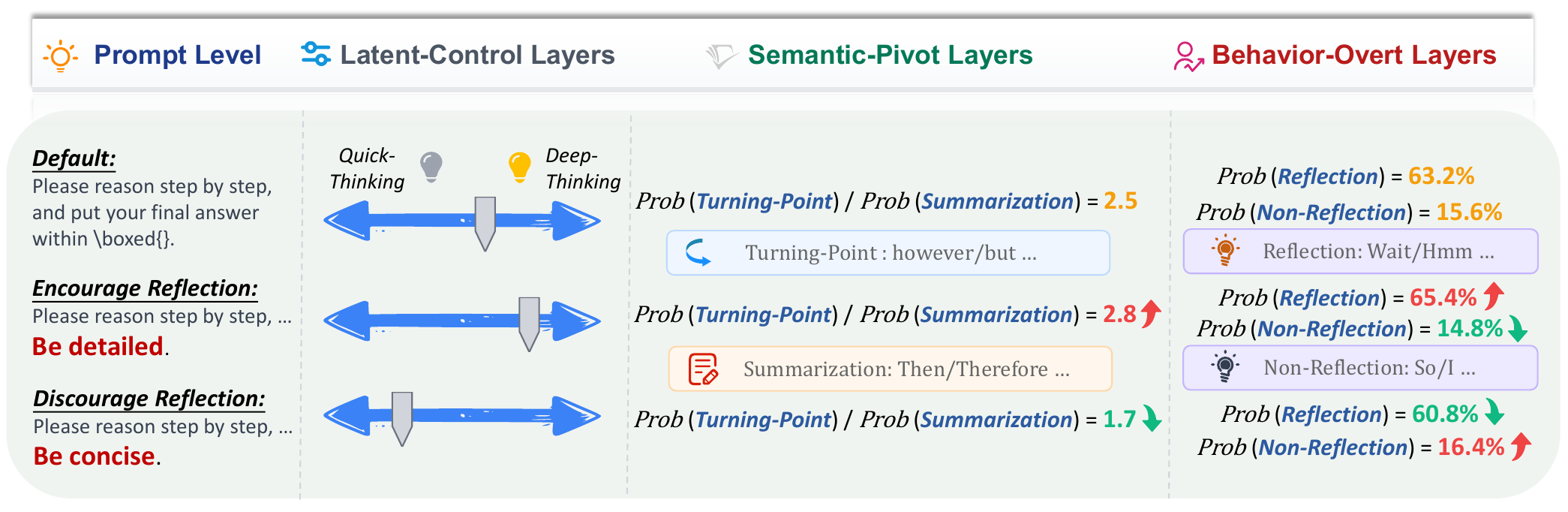}
  \caption{\textbf{Depth-wise causal chain of reflection.} Prompt-level cues modulate latent-control activations toward a deep or quick-thinking direction, shift probability mass between turning-point and summarization cues in semantic-pivot layers, and ultimately change the sampling likelihood of reflection markers in behavior-overt layers.}
  \label{fig_intro}
  \vspace{-0.6cm}
\end{figure*}

In our study, we adopt the logit lens as a probe, applying the output decoder to intermediate-layer activations to read out token-level semantics~\cite{nostalgebraist2020logit_lens,belrose2023eliciting,fang2025unsupervised}. 
Based on this, we conduct a correlation analysis to characterize the functional roles of different depths from two complementary views. 
First, we decode contrastive activations between paired prompts that explicitly encourage versus discourage thinking, which helps isolate latent control signals that may be obscured in raw activations~\cite{farooq2025sentiment,arditi2024refusal}. 
This reveals a contiguous block of early-to-mid layers with clear thinking-budget semantics (e.g., detailed/concise), which we term latent-control layers. 
Second, we anchor on the emergence of the reflection marker token ``Wait'' and trace activations along the model's natural forward pass. 
In later intermediate layers, probability mass shifts sharply toward discourse cues, including turning-point tokens (e.g., but/however) and summarization tokens (e.g., so/therefore), forming semantic-pivot layers.
And in the last layers, reflection-behavior tokens (e.g., Wait) become increasingly likely until they are sampled, defining behavior-overt layers. 
Together, these views suggest a stage-wise organization of reflection behavior across depth.

To further validate this structured progression, we hypothesize a depth-wise causal chain: prompt-level control $\rightarrow$ latent-control layers $\rightarrow$ semantic-pivot layers $\rightarrow$ behavior-overt layers. 
We test this hypothesis via targeted interventions, either (i) injecting explicit semantics at the prompt level or (ii) applying activation steering within the latent-control layers, and then measuring downstream effects. 
For instance, as illustrated in Fig.~\ref{fig_intro}, the intervention of discouraging reflection shifts activations along the latent-control direction toward a quick-thinking budget.
This perturbation propagates to later stages: in the semantic-pivot layers, probability mass moves from turning-point tokens to summarization tokens. 
And in the behavior-overt layers, the sampling likelihood of reflection-behavior tokens decreases. 
In contrast, encouraging reflection will yield the opposite trends.
Moreover, fine-grained activation-steering experiments exhibit the same qualitative propagation pattern, further supporting a coherent depth-wise causal chain.

Our results suggest that the internal propagation process underlying reflection follows a human-like meta-cognitive progression—moving from latent monitoring, to discourse-level regulation, and finally to overt self-reflection. 
Moreover, we anchor on additional reflection markers and draw on data from diverse domains, showing that this stage-wise pattern can generalize broadly. 
Overall, our contributions can be summarized as follows:
\begin{itemize}[leftmargin=*,noitemsep,topsep=0pt]
\item \textbf{Mechanistic Decomposition.} We present a depth-wise mechanistic account of reflection in R1-style LLMs by identifying latent-control, semantic-pivot, and behavior-overt stages. This links internal representations to discourse cues and explicit reflection markers.
\item \textbf{Causal Verification.} We validate a coherent causal chain across these stages using prompt-level semantic and activation steering interventions. 
The induced effects propagate stage-by-stage, ultimately modulating the sampling likelihood of reflection markers.
\item \textbf{Robust Generalization.} To the best of our knowledge, we are the first to characterize the depth-wise mechanisms underlying reflective behaviors. Our analysis generalizes across models, anchored tokens, and datasets, providing a foundation for future work on controllable reflection.
\end{itemize}

\vspace{-0.2cm}
\section{Preliminary}~\label{prelimi}
\vspace{-0.9cm}

\subsection{Logit-Lens Decoding}
Let $f_\theta$ be an autoregressive Transformer with $L$ layers, hidden size $d$, and vocabulary $\mathcal{V}$. Given a prefix $x_{1:t}$, let $h_t^{(\ell)}\in\mathbb{R}^d$ denote the residual-stream representation at position $t$ after layer $\ell$. Let $W_U\in\mathbb{R}^{|\mathcal{V}|\times d}$ be the unembedding matrix and $\mathrm{LN}(\cdot)$ the final layer normalization before the output head. We use the logit lens as a generic probe by reusing this head to decode any vector $v\in\mathbb{R}^d$:
$
z(v) = W_U\,\mathrm{LN}(v), 
p(v) = \mathrm{softmax}\!\big(z(v)\big).
$
This probe guides two complementary views.
\textbf{(i) Contrastive activation-difference decoding.}
For a paired prompt set $(x^{+},x^{-})$ that differs only in a targeted semantic attribute, we compute the layer-wise activation difference at the same depth and final token position,
$
\Delta h_{-1}^{(\ell)} = h_{-1}^{(\ell)}(x^{+}) - h_{-1}^{(\ell)}(x^{-}),
$
and decode it with the logit lens, achieving $p(\Delta h_{-1}^{(\ell)})$.
Decoding $\Delta h_{-1}^{(\ell)}$ highlights directions most associated with the controlled attribute, reducing confounds present in raw activations.
\textbf{(ii) Layer-wise forward-activation decoding.}
Setting $h_t^{(\ell)}$ yields layer-indexed next-token logits and distributions. 
We apply this probe to achieve $p(h_t^{(\ell)}$ along the model’s natural forward propagation at reflection onset: we align $t$ to the position immediately before a reflection marker (e.g., ``Wait'') is emitted, and track the decoded signals across layers. 
This anchored view reveals which token-level semantics become linearly readable at depth $\ell$ when reflection emerges.

\vspace{-0.2cm}
\subsection{Experiment Settings}\label{sec_exp_set}

\paragraph{LLMs.}

We study two representative R1-style LLMs released by different organizations: DeepSeek-R1-Distill-Qwen-7B (DeepSeek-R1$_{7B}$)~\cite{guo2025deepseek} and Qwen3-4B-Thinking-2507 (Qwen3-Think$_{4B}$)~\cite{yang2025qwen3}. 
DeepSeek-R1$_{7B}$, released by \texttt{deepseek-ai}, is a distilled dense checkpoint trained on reasoning traces from DeepSeek-R1$_{671B}$, and follows a Qwen2-style decoder-only Transformer with 28 layers and grouped-query attention (28 query heads, 4 KV heads). 
Qwen3-Think$_{4B}$, released by Alibaba's Qwen team, adopts a deeper 36-layer architecture with grouped-query attention (32 query heads, 8 KV heads). 
Together, these models form complementary testbeds to assess whether the structured progression we identify generalizes across implementations and scales.

\paragraph{Analysis Data.}
We use GSM8K~\cite{cobbe2021training} as our primary benchmark and sample 200 problems for experiments.
To obtain contrastive activation differences, we construct paired prompts $(x^{+},x^{-})$ by appending to each instruction either (i) a reflection-encouraging suffix (e.g., ``Your response must include a detailed reasoning process.'') or (ii) a reflection-suppressing suffix (e.g., ``Your response must include a concise reasoning process.''), yielding 200 matched pairs. 
For forward-activation analyses, we run greedy decoding under a default prompt and record model outputs. 
Following Sec.~\ref{prelimi}, we probe intermediate-layer activations only at reflection onset by aligning to the emission of the marker token ``Wait''. 
Each `Wait'' occurrence is treated as one anchored sample, resulting in 535 reflection-onset samples for DeepSeek-R1$_{7B}$ and 1553 for Qwen3-Think$_{4B}$.
\fbox{\parbox{\columnwidth}{%
\textbf{An anchored sample.} \texttt{\textless|begin\_of\_sentence| \textgreater\textless|User|\textgreater} Twenty dozen cups cost \$1200 less than ... Calculate the total cost of buying each cup. \texttt{\textless|Assistant|\textgreater} \texttt{\textless think\textgreater} Okay, so I have this problem here: ... I need to find the total cost of each cup. \textbf{\emph{Wait}}
}}

\section{Correlation Analysis}\label{corr_analy}

\begin{figure*}[ht]
  \centering
  \begin{subfigure}[t]{0.9\textwidth}
    \centering
    \includegraphics[width=\linewidth]{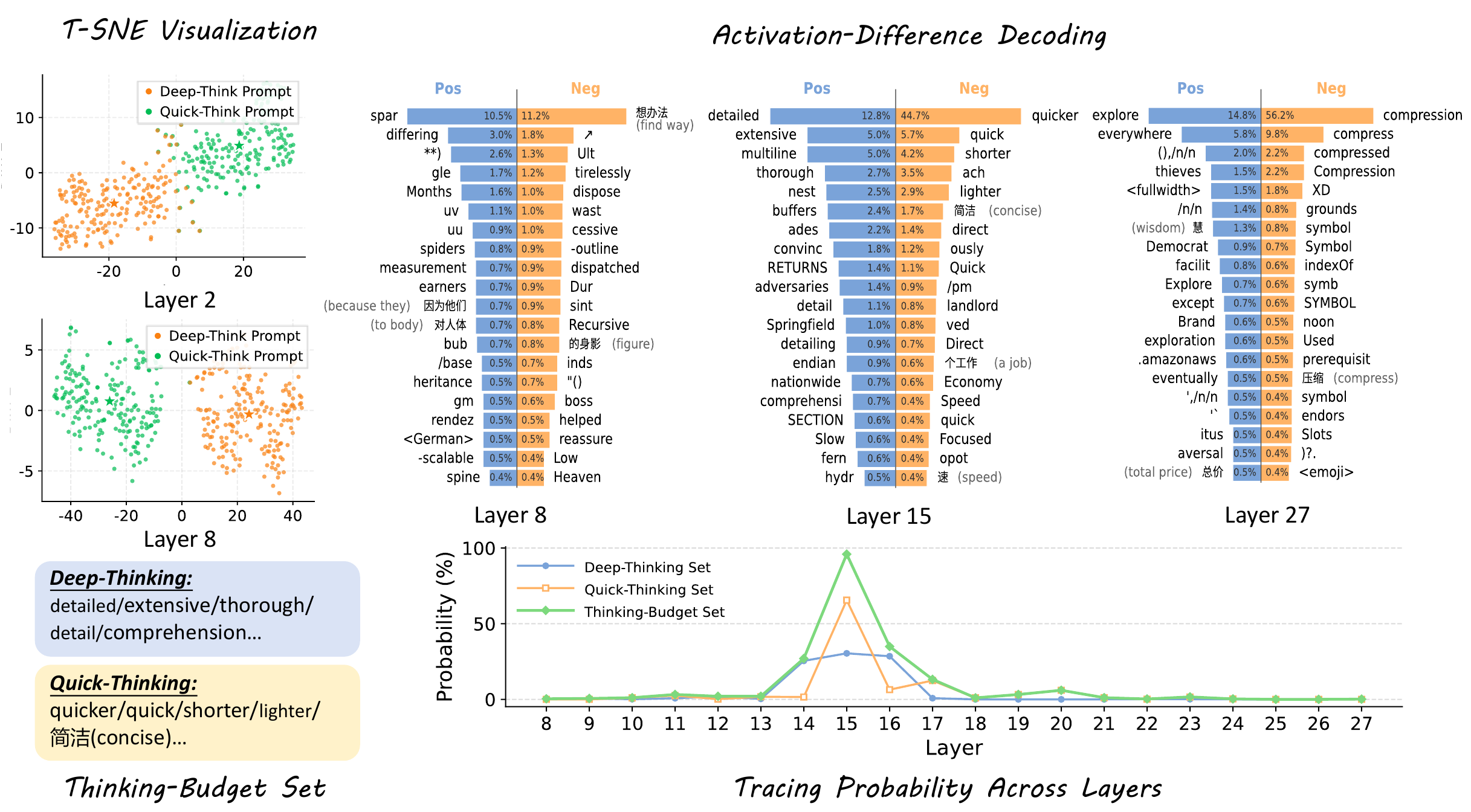}
    \caption{Analysis of Activation-Difference. The asterisk (*) next to a token denotes a placeholder for a whitespace character.}
    \label{fig:dsqw_diff}
  \end{subfigure}\hfill
  \begin{subfigure}[t]{0.92\textwidth}
    \centering
    \includegraphics[width=\linewidth]{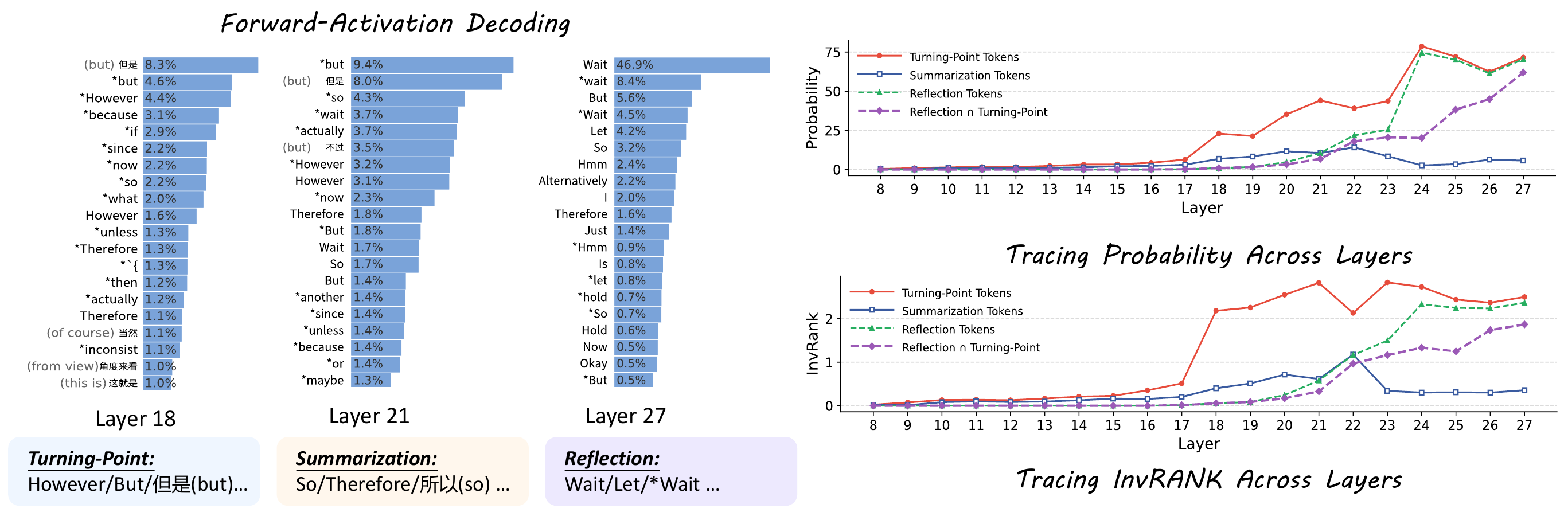}
    \caption{Analysis of Forward-Activation. The asterisk (*) next to a token denotes a placeholder for a whitespace character.}
    \label{fig:dsqw_forward}
  \end{subfigure}
  \caption{Analysis on DeepSeek-R1$_{7B}$. We report analyses of activation-difference and forward-activation, including t-SNE visualizations, logit-lens decoding results, and our predefined sets. The layer-wise trajectories highlight that different layers play distinct functional roles.
}
  \vspace{-0.5cm}
  \label{fig_dsqw_ana}
\end{figure*}


\subsection{Activation-Difference Analysis}\label{sec_acti_diff_ana}

\begin{figure*}[ht]
  \centering
  \begin{subfigure}[t]{0.9\textwidth}
    \centering
    \includegraphics[width=\linewidth]{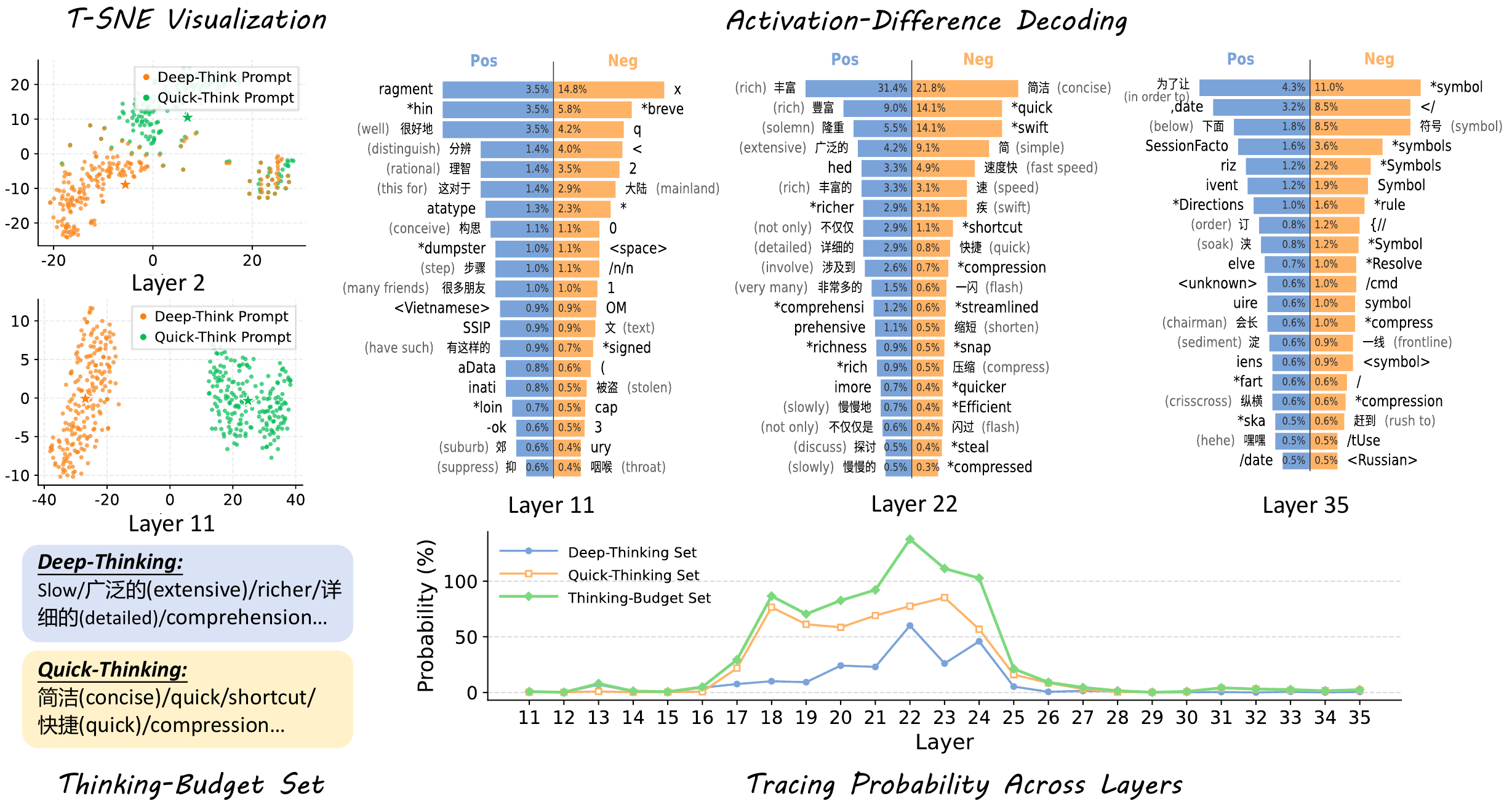}
    \caption{Analysis of Activation-Difference. The asterisk (*) next to a token denotes a placeholder for a whitespace character.}
    \label{fig:qw3_diff}
  \end{subfigure}\hfill
  \begin{subfigure}[t]{0.92\textwidth}
    \centering
    \includegraphics[width=\linewidth]{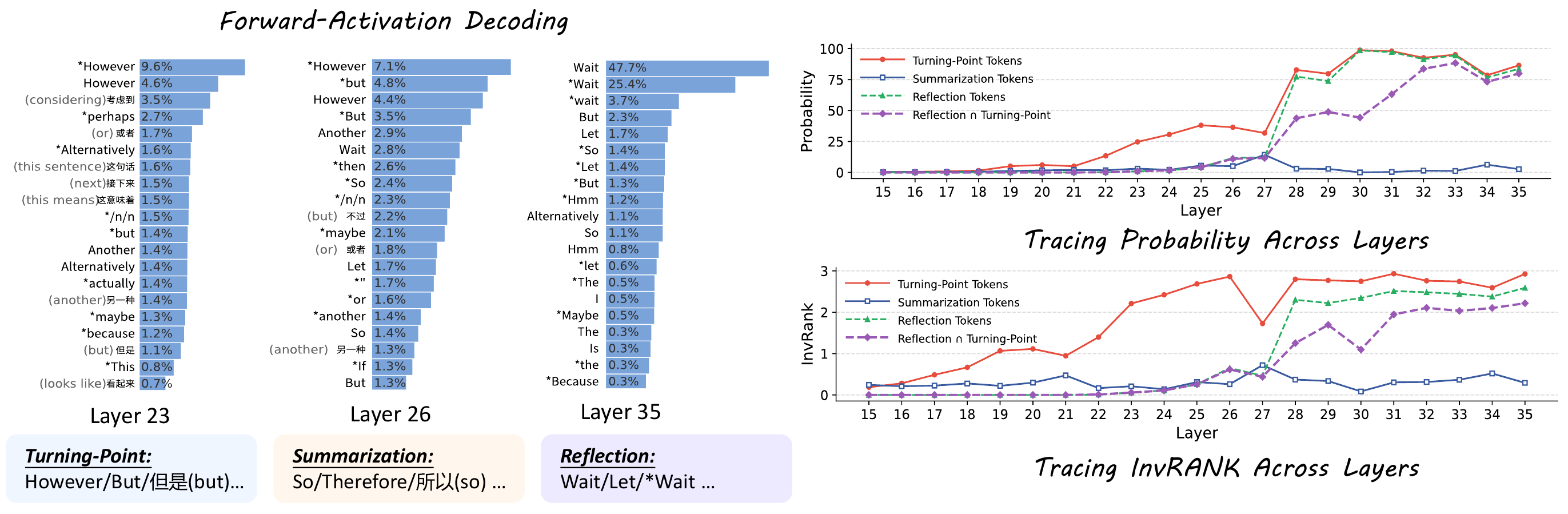}
    \caption{Analysis of Forward-Activation. The asterisk (*) next to a token denotes a placeholder for a whitespace character.}
    \label{fig:qw3_forward}
  \end{subfigure}
  \caption{Analysis on Qwen3-Think$_{4B}$. We report analyses of activation-difference and forward-activation, including t-SNE visualizations, logit-lens decoding results, and our predefined sets. The layer-wise trajectories highlight that different layers play distinct functional roles.}
\vspace{-0.5cm}
  \label{fig_qw3_ana}
\end{figure*}

Before decoding activation differences, we visualize layer-wise representations of contrastive prompt pairs with T-SNE~\cite{maaten2008visualizing}. 
As shown in the upper-left panels of Figs.~\ref{fig:dsqw_diff} and \ref{fig:qw3_diff}, the two prompts become clearly separable in early-to-mid layers, around Layer 8 for DeepSeek-R1$_{7B}$ and Layer 11 for Qwen3-Think$_{4B}$, and this separation largely persists to the final layer (see App.~\ref{app_visual_tsne} for additional plots). 
This progressive, near-linear separation suggests that thinking-related semantics may be organized along an approximately linear direction in representation space.
Building on this, we probe activation differences with the logit-lens to interpret their token-level semantics. 
We define the averaged difference induced by $x^{+}-x^{-}$ as the positive direction $d_{\text{pos}}$, and $x^{-}-x^{+}$ as the negative direction $d_{\text{neg}}$. 
As shown in the upper-right panels of Figs.~\ref{fig:dsqw_diff} and \ref{fig:qw3_diff}, there exists a mid-layer region where decoded tokens become especially salient for thinking-budget semantics (Layer 15 for DeepSeek-R1$_{7B}$ and Layer 22 for Qwen3-Think$_{4B}$, and more examples in App.~\ref{app_visual_diff_decode}). 
Interestingly, decoding $d_{\text{pos}}$ surfaces deep-thinking cues (e.g., detailed/extensive), while decoding $d_{\text{neg}}$ yields the opposite semantics (e.g., concise/shorter). 
Moreover, despite GSM8K being in English, these layers may also decode semantically meaningful Chinese tokens, echoing the intuition that internal deliberation can surface in a familiar language.

To quantify the effect, we collect deep and quick-thinking tokens from Top-50 decoded tokens to form a thinking-budget set, and track their decoded probability mass across depth (as shown in the bottom panels of Figs.~\ref{fig:dsqw_diff} and \ref{fig:qw3_diff}). 
We observe a monotonic rise over a contiguous interval, most pronounced over Layers 13-15 for DeepSeek-R1$_{7B}$ and Layers 16-22 for Qwen3-Think$_{4B}$. 
Beyond these ranges, deeper layers no longer produce comparably salient thinking-budget decoding, suggesting a shift of representational capacity toward higher-level semantics.
Based on these observations, we define the \textbf{latent-control layers} as the interval from the first layer where contrastive prompts become separable to the layer where thinking-budget semantics are maximally dominant under logit-lens decoding: Layers 8-15 for DeepSeek-R1$_{7B}$ and Layers 11-22 for Qwen3-Think$_{4B}$. 
In Sec.~\ref{sec_ana_act_steer}, we further show that intervening within this interval often yields strong linear control over thinking length while minimizing degradation in language quality.

\subsection{Forward-Activation Analysis}\label{sec:forw_acti_ana}

Our study further traces the decoded states of intermediate-layer activations at the moment when the reflection token ``Wait'' emerges.
Using the samples collected in Sec.~\ref{sec_exp_set}, we apply the logit-lens at each layer $\ell$ and compute the mean decoded probability $p_{\ell}(t)$ for token $t$ (averaged over all anchored samples), together with its layer-wise rank $\mathrm{rank}_{\ell}(t)$. 
As shown in Figs.~\ref{fig:dsqw_forward} and \ref{fig:qw3_forward} (Top-20 token visualizations for some layers and full results in App.~\ref{app_visual_forward_decode}), we observe a local stage-wise pattern: 
in specific intermediate layers, discourse-level pivot cues, including turning-point tokens (e.g., but/however) and summarization tokens (e.g., so/therefore), rise sharply to high ranks. 
For DeepSeek-R1$_{7B}$, this pivoting becomes salient around Layer 18 and for Qwen3-Think$_{4B}$, around Layer 23. 
Reflection-behavior tokens (e.g., Wait), however, are not yet dominant in these pivot layers. 
Their probabilities increase markedly only in the subsequent layers (e.g. ``Wait'' begins to enter the Top-20 at Layer 21 for DeepSeek-R1$_{7B}$ and Layer 26 for Qwen3-Think$_{4B}$) and continue rising toward the final layer until they become highly sampleable. 
These observations suggest a progression in which the model first constructs discourse-level pivot cues, which precede the overt emergence of reflection markers.

To quantify this progression, we define three token sets: a turning-point set $\mathcal{T}$, a summarization set $\mathcal{S}$, and a reflection-behavior set $\mathcal{R}$. 
We first identify the empirically most salient pivot layer $\ell^\star$ and collect its Top-50 decoded tokens, and then form $\mathcal{T}$ and $\mathcal{S}$ by intersecting them with turning-point and summarization markers from PDTB~\cite{prasad2017penn} and its Chinese counterpart (CDTB), respectively. 
For $\mathcal{R}$, we collect reflection-behavior tokens from the Top-10 decoded tokens at the final layer. 
All defined token sets can be found in App.~\ref{app_pre_difine_tokens}.
For any set $\mathcal{A}$, we track (i) the overall probability mass
$
P_{\ell}(\mathcal{A}) \;=\; \sum_{t \in \mathcal{A}} p_{\ell}(t)
$
and (ii) a rank-based score  by summing the reciprocal of ranks:
\begin{equation}
\mathrm{InvRANK}_{\ell}(\mathcal{A}) \;=\; \left(\sum_{t \in \mathcal{A}} \mathrm{rank}_{\ell}(t)^{-1}\right)
\end{equation}
where larger values indicate higher overall ranking. 
Since some reflection tokens can occasionally appear among the Top-50 tokens at $\ell^\star$ (and thus enter $\mathcal{T}$), we additionally report the overlap set $\mathcal{I}=\mathcal{T}\cap\mathcal{R}$ to control for this confound.

The right panels of Figs.~\ref{fig:dsqw_forward} and \ref{fig:qw3_forward} summarize the layer-wise trajectories. 
For $\mathcal{T}$ and $\mathcal{S}$, both $P_{\ell}(\cdot)$ and $\mathrm{InvRANK}_{\ell}(\cdot)$ begin to rise from an intermediate layer $m$ onward. After a later transition around layer $k$, $\mathcal{S}$ declines while $\mathcal{T}$ remains relatively high. 
Notably, the sustained high level of $\mathcal{T}$ in later layers is partly driven by its overlap with reflection-behavior tokens: 
beyond layer $k$, the overlap set $\mathcal{I}$ rises sharply and accounts for a substantial fraction of $\mathcal{T}$, indicating that the driver of high $\mathcal{T}$ shifts from semantic-pivot cues to reflection-behavior tokens in late layers.
Moreover, $\mathcal{R}$ exhibits a sharp increase in both mass and rank near the same region (around layer $k$), consistent with a staged transition from discourse-level pivoting to overt reflection. 
Motivated by these trajectories, we introduce an explicit layer-wise partition. 
We define layers $m$-$k$ as the \textbf{semantic-pivot layers}, where discourse cues dominate the decoded distributions, and layers $k$ through the final layer as the \textbf{behavior-overt layers}, where reflection tokens rapidly rise until they become highly likely to be sampled. 
Concretely, for DeepSeek-R1$_{7B}$, we identify Layers 16-21 as semantic-pivot layers and Layers 22-27 as behavior-overt layers, and for Qwen3-Think$_{4B}$, Layers 23-27 constitute the semantic-pivot layers and Layers 28-35 the behavior-overt layers.

\section{Causal Intervention Analysis}\label{cau_analy}

In Sec.~\ref{corr_analy}, logit-lens probing reveals qualitatively different semantic signals across depth, motivating a three-stage partition of the forward propagation: \textbf{latent-control layers}, \textbf{semantic-pivot layers}, and \textbf{behavior-overt layers}. 
In latent-control layers, we identify an approximately linear residual direction that modulates the model's thinking budget.
In semantic-pivot layers, discourse cues with clear semantic roles (e.g., turning-point and summarization markers) emerge; 
And in behavior-overt layers, explicit reflection markers become increasingly probable until they are sampled.
Building on these correlational observations, we test a stronger causal hypothesis: \emph{prompt} $\rightarrow$ \emph{latent-control} $\rightarrow$ \emph{semantic-pivot} $\rightarrow$ \emph{behavior-overt}. 
We conduct targeted interventions at two granularities: (i) prompt-level semantic manipulation and (ii) activation steering in latent-control layers, and examine how their effects propagate to downstream stages. 
Throughout, we use the same dataset as in Sec.~\ref{sec_exp_set}, anchoring on the occurrence of the reflection token ``Wait'' to track stage-wise changes under intervention.


\begin{figure}[t]
  \centering
  \begin{subfigure}[t]{0.24\textwidth}
    \centering
    \includegraphics[width=\linewidth]{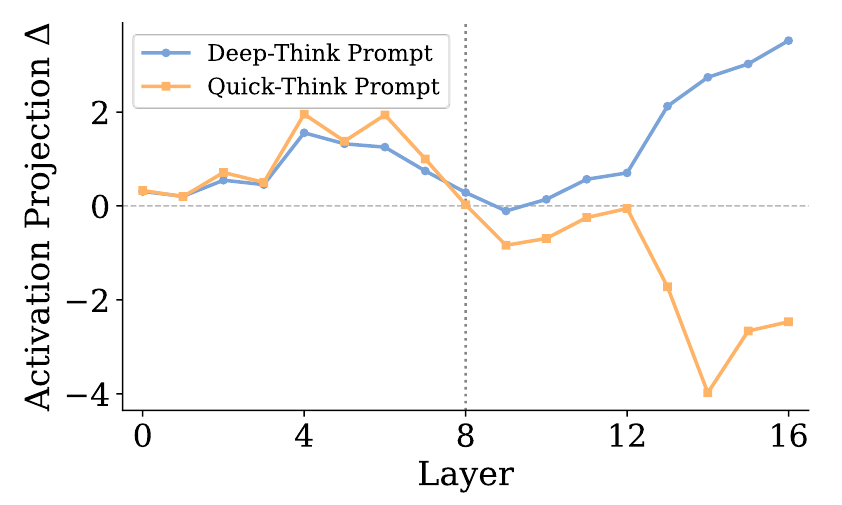}
    \caption{Analysis on DeepSeek-R1$_{7B}$.}
  \end{subfigure}\hfill
  \begin{subfigure}[t]{0.24\textwidth}
    \centering
    \includegraphics[width=\linewidth]{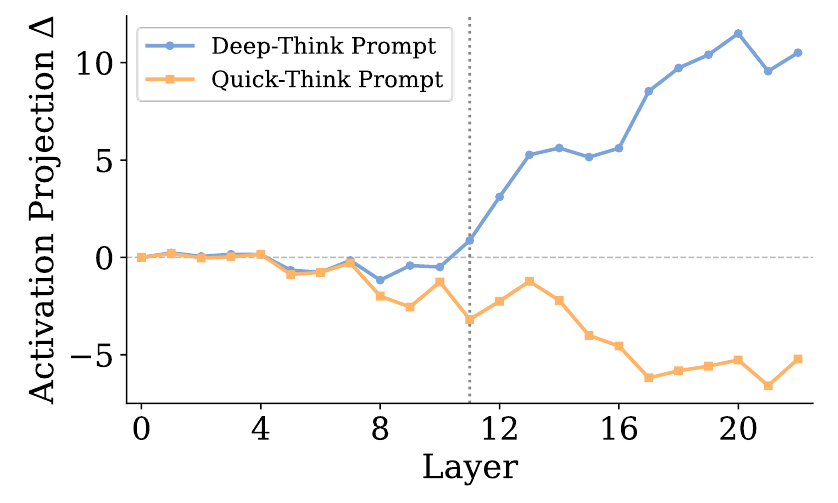}
    \caption{Analysis on Qwen3-Think$_{4B}$.}
  \end{subfigure}
  \caption{Changes of projection strength in latent-control layers under prompt-level interventions.}
        \vspace{-0.3cm}
  \label{fig:proj_analy}
\end{figure}


\subsection{Intervention Settings}

\paragraph{Prompt-level intervention.}
As illustrated in Fig.~\ref{fig_intro}, we perform a minimal semantic manipulation by appending either a deep-thinking cue (``Be detailed'') or a quick-thinking cue (``Be concise'') to the default prompt, aiming to encourage or suppress reflective behavior, respectively.

\paragraph{Activation-steering intervention.}
We additionally intervene internally by applying activation steering within the latent-control layers, using the direction $\mathbf{d}^{(l)}_{\text{pos}}$ extracted in Sec.~\ref{sec_acti_diff_ana}.
For an input $x$, we modify the residual-stream state at token position $i$ in layer $l$ as
$
\mathbf{h}^{(l)}_{i}(x) \leftarrow \mathbf{h}^{(l)}_{i}(x) + \alpha \cdot \mathbf{d}^{(l)}_{\text{pos}},
$
where $\alpha$ controls the intervention strength: $\alpha>0$ steers activations toward a deep-thinking regime (encouraging reflection), while $\alpha<0$ steers toward a quick-thinking regime (suppressing reflection). 
To avoid language degradation under overly large $|\alpha|$, we restrict $\alpha$ to a stable range based on the calibration in Sec.~\ref{sec_ana_act_steer}.



\subsection{Effect on Latent-Control Layers}

Since activation-steering has operated on the latent-control layers, we only analyze the effect of prompt-level intervention on the latent-control layers.
Concretely, we quantify how a prompt modification changes the projection of activations onto the thinking-budget direction $\mathbf{d}^{(l)}_{\text{pos}}$ by the following metric $\Delta^{(l)}$:
\begin{equation}
\begin{aligned}
s^{(l)}_{i}(x) &= \big\langle \mathbf{h}^{(l)}_{i}(x), \mathbf{d}^{(l)}_{\text{pos}} \big\rangle,\\
\Delta^{(l)} &= \mathbb{E}_{x \sim \mathcal{D}}\;\mathbb{E}_{i \in \mathcal{M}(x)}
\Big[\, s^{(l)}_{i}(x_{\text{int}}) - s^{(l)}_{i}(x_{\text{base}})\,\Big],
\end{aligned}
\end{equation}
where $\mathbf{h}$ denotes the hidden state, $x_{\text{base}}$ the default prompt, $x_{\text{int}}$ the intervened prompt, and $\mathcal{M}(x)$ represents the anchored token position.
As shown in Fig.~\ref{fig:proj_analy}, we report $\Delta^{(l)}$ for the latent-control layers as well as earlier layers. 
In early layers, we observe no consistent trend. 
In contrast, within our identified latent-control layers, prompt semantics induce a clear directional shift: adding a deep-thinking cue yields $\Delta^{(l)}\!>\!0$ (steering toward a deep-thinking budget), whereas adding a quick-thinking cue yields $\Delta^{(l)}\!<\!0$. 
This pattern supports that latent-control layers serve as a dedicated control stage where prompt-level semantics are reliably translated into a thinking-budget signal.

\begin{figure}[t]
  \centering
  \begin{subfigure}[t]{0.24\textwidth}
    \centering
    \includegraphics[width=\linewidth]{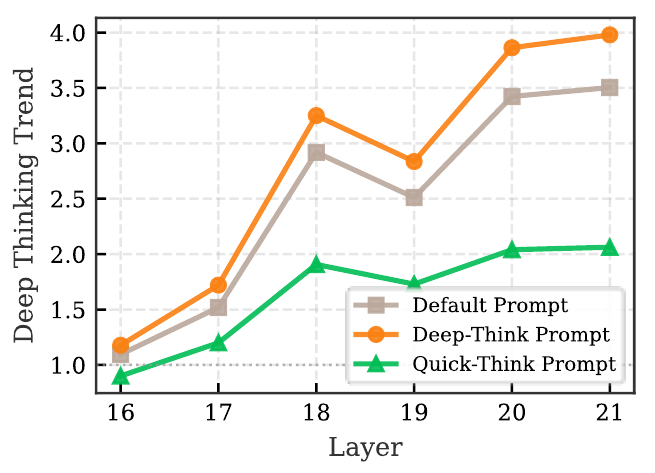}
    \caption{Analysis on DeepSeek-R1$_{7B}$.}
  \end{subfigure}\hfill
  \begin{subfigure}[t]{0.24\textwidth}
    \centering
    \includegraphics[width=\linewidth]{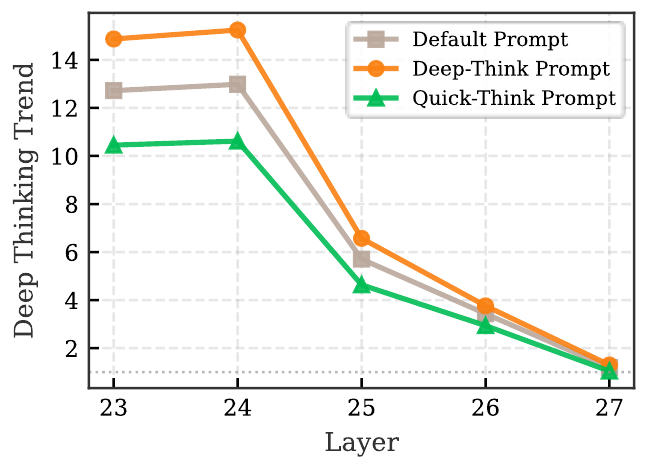}
    \caption{Analysis on Qwen3-Think$_{4B}$.}
  \end{subfigure}
  \caption{Effect on pivot-semantic layers under prompt-level interventions.}
      \vspace{-0.3cm}
  \label{fig:piv_sem_prompt_ana}
\end{figure}

\begin{figure}[t]
  \centering
  \begin{subfigure}[t]{0.24\textwidth}
    \centering
    \includegraphics[width=\linewidth]{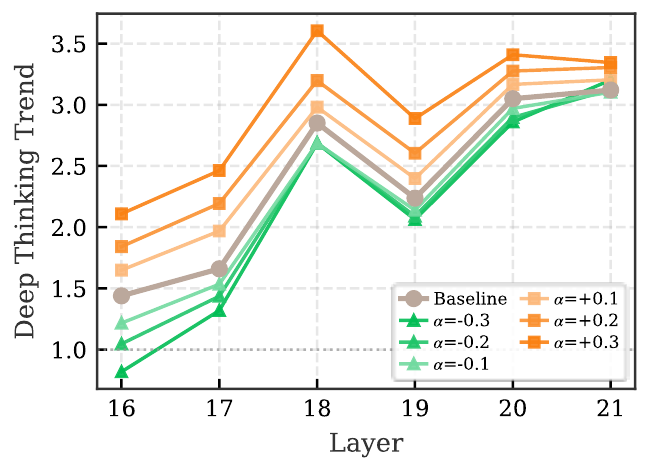}
    \caption{Analysis on DeepSeek-R1$_{7B}$.}
  \end{subfigure}\hfill
  \begin{subfigure}[t]{0.24\textwidth}
    \centering
    \includegraphics[width=\linewidth]{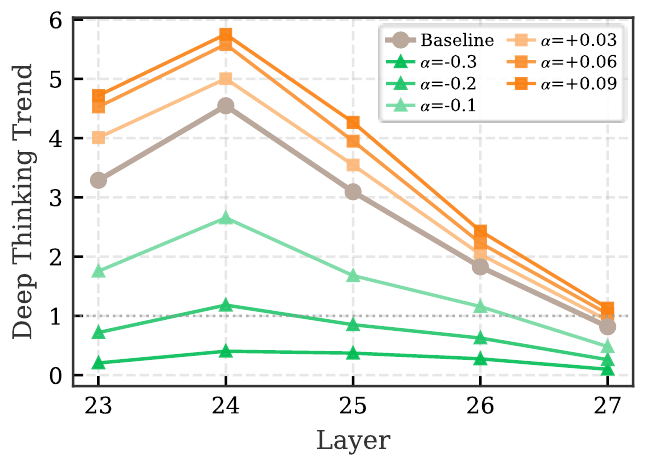}
    \caption{Analysis on Qwen3-Think$_{4B}$.}
  \end{subfigure}
  \caption{Effect on pivot-semantic layers under activation-steering interventions.}
      \vspace{-0.3cm}
  \label{fig:piv_sem_steer_ana}
\end{figure}

\begin{figure}[ht]
  \centering
  \begin{subfigure}[t]{0.39\textwidth}
    \centering
    \includegraphics[width=\linewidth]{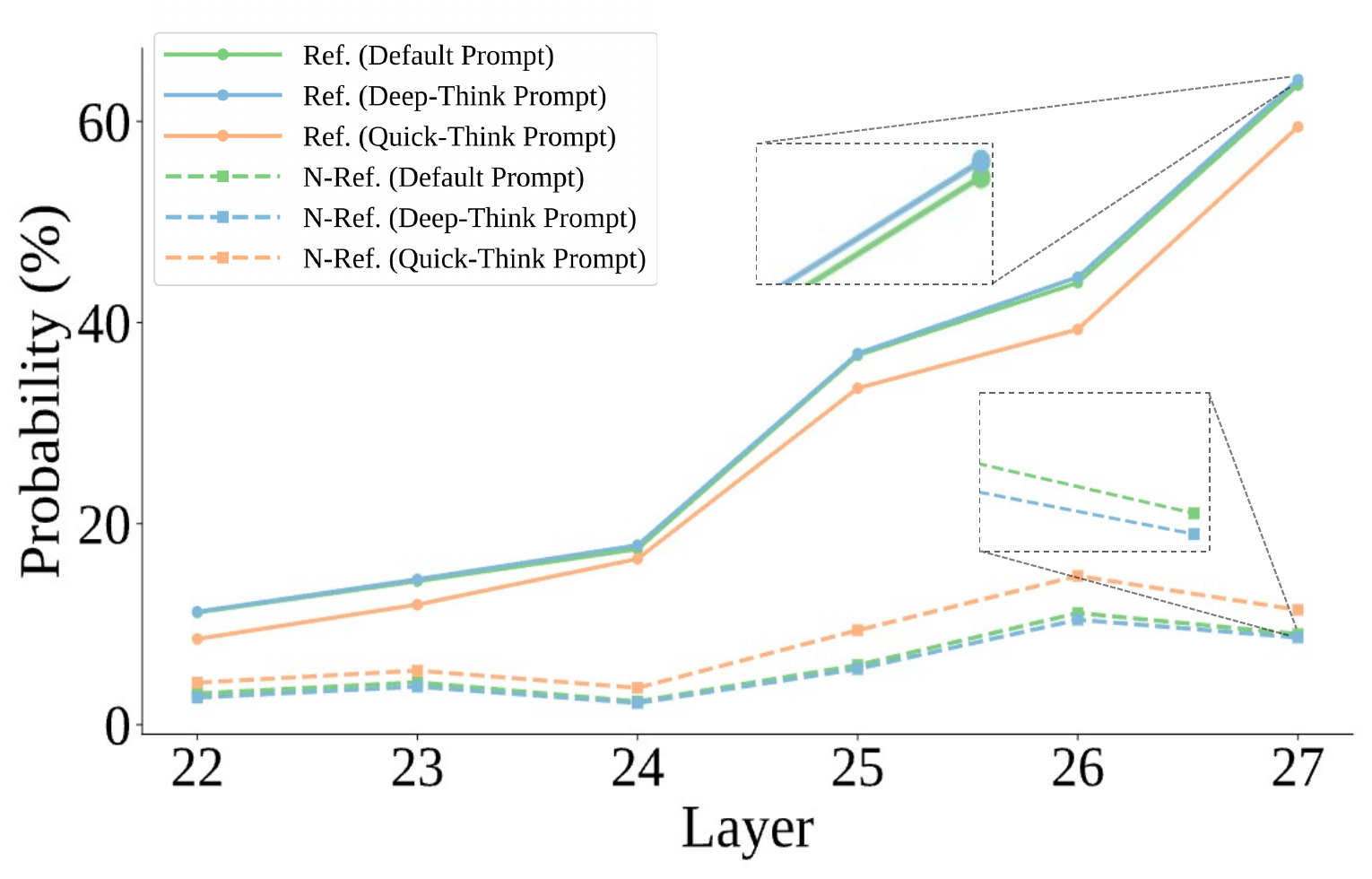}
    \caption{Analysis on DeepSeek-R1$_{7B}$.}
  \end{subfigure}\hfill
  \begin{subfigure}[t]{0.39\textwidth}
    \centering
    \includegraphics[width=\linewidth]{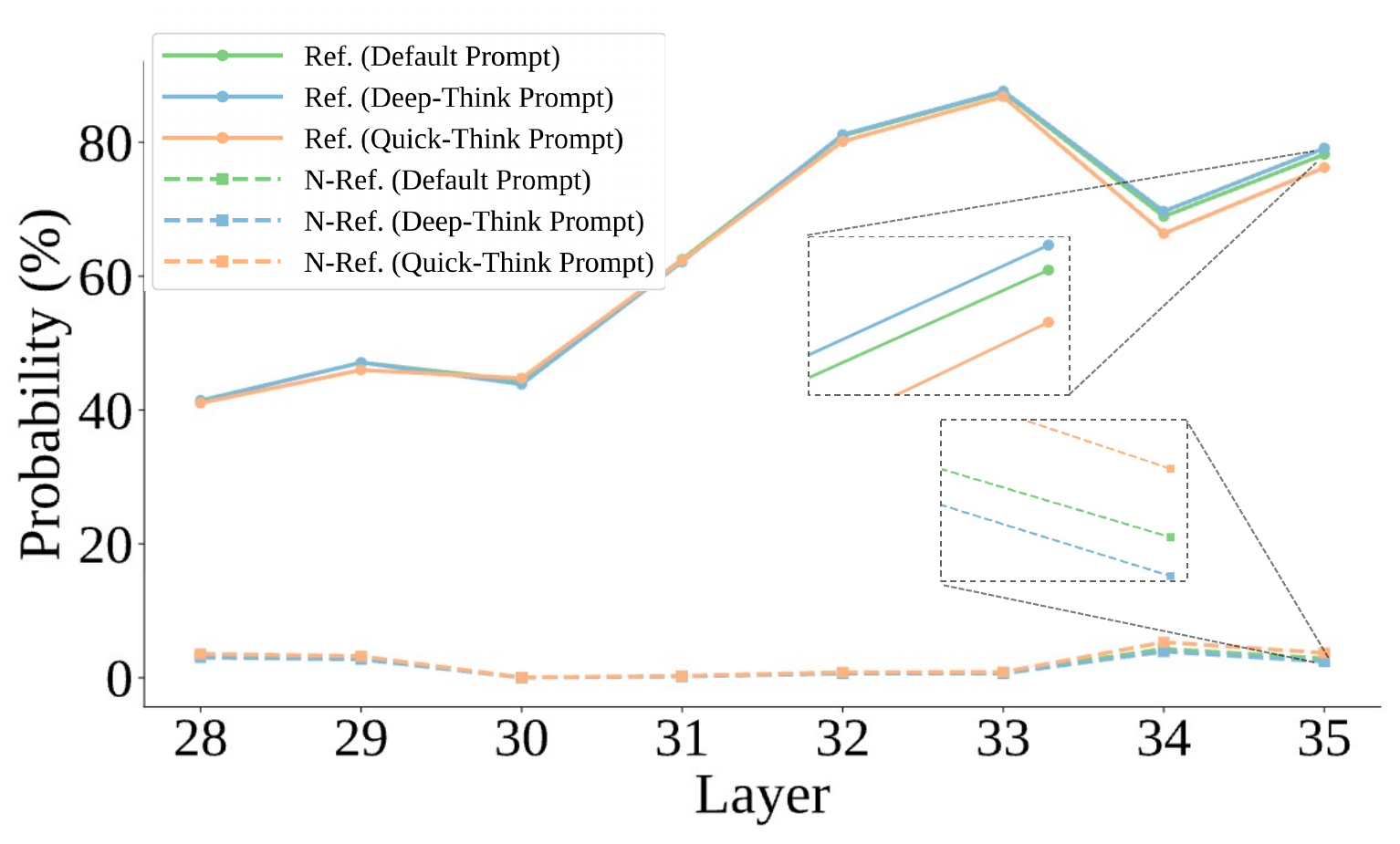}
    \caption{Analysis on Qwen3-Think$_{4B}$.}
  \end{subfigure}
  \caption{Effect on behavior-overt layers under prompt-level interventions. Ref. and N-Ref. represent $\mathcal{R}$ and $\mathcal{NR}$ sets.}
      \vspace{-0.3cm}
\label{fig:behvave_prompt_ana}
\end{figure}

\begin{figure}[ht]
  \centering
  \begin{subfigure}[t]{0.45\textwidth}
    \centering
    \includegraphics[width=\linewidth]{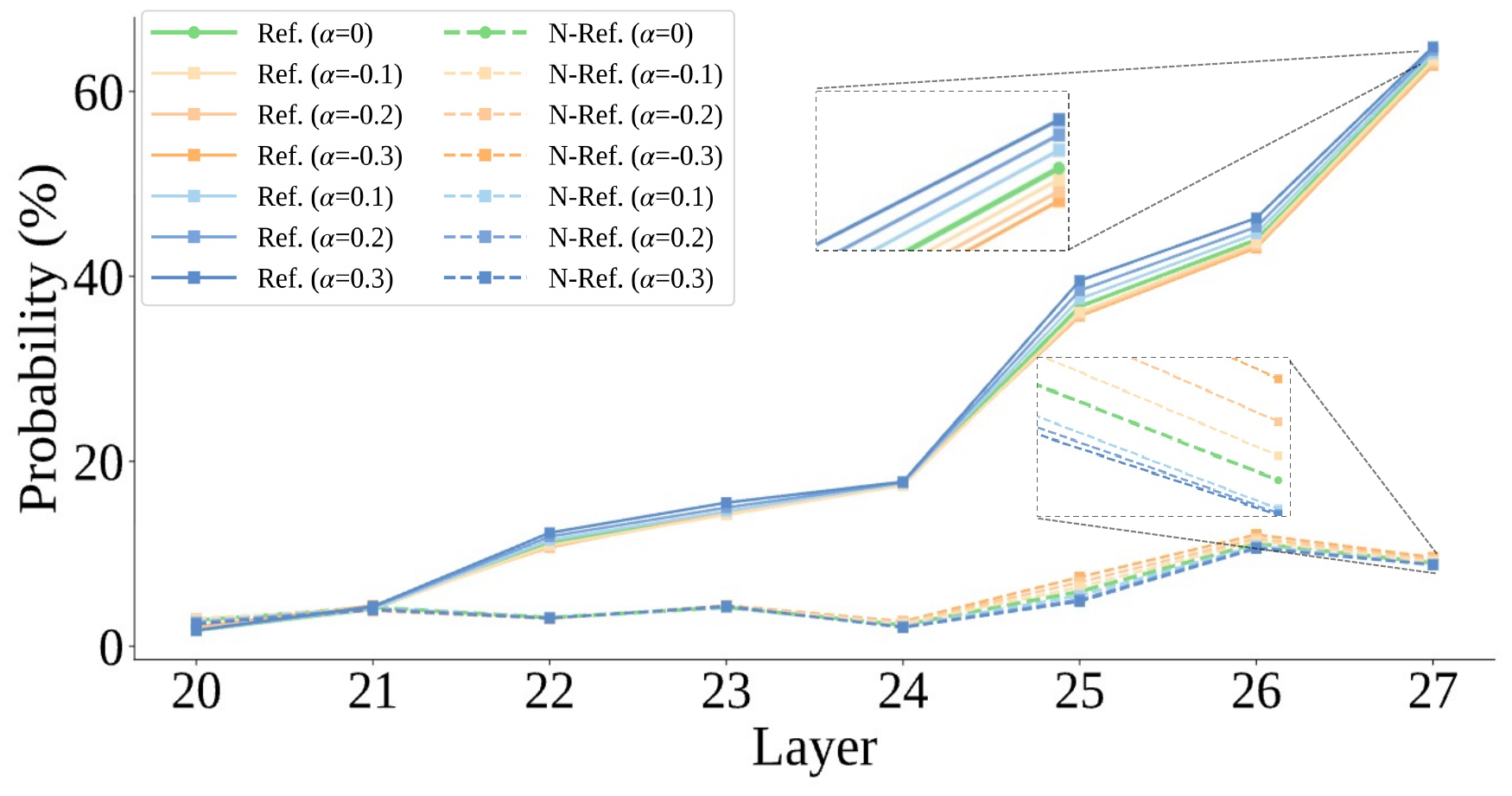}
    \caption{Analysis on DeepSeek-R1$_{7B}$.}
  \end{subfigure}\hfill
  \begin{subfigure}[t]{0.45\textwidth}
    \centering
    \includegraphics[width=\linewidth]{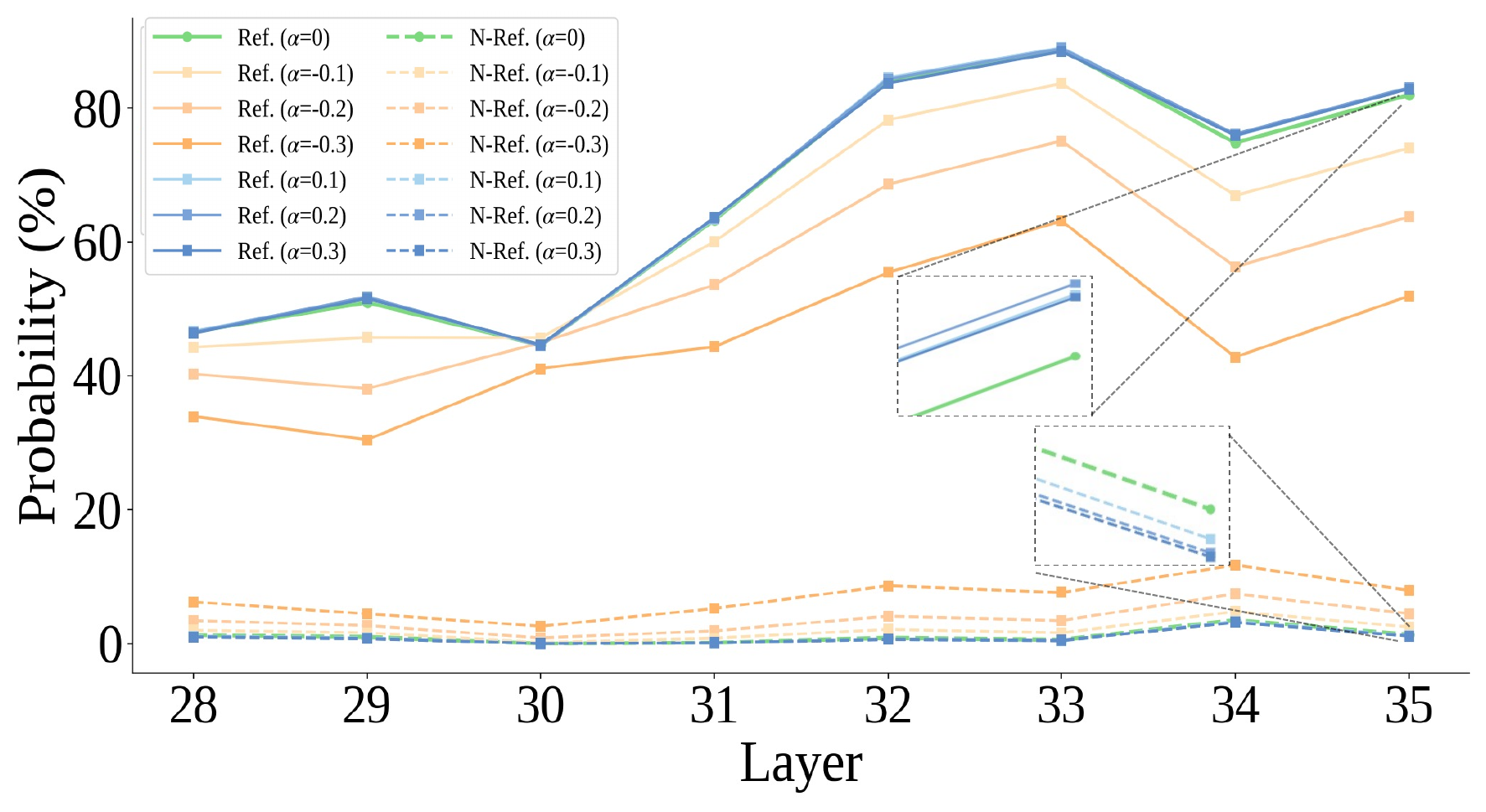}
    \caption{Analysis on Qwen3-Think$_{4B}$.}
  \end{subfigure}
  \caption{Effect on behavior-overt layers under activation-steering interventions. Ref. and N-Ref. represent $\mathcal{R}$ and $\mathcal{NR}$ sets.}
      \vspace{-0.3cm}
\label{fig:behvave_steer_ana}
\end{figure}

\subsection{Effect on Semantic-Pivot Layers}

Within the semantic-pivot layers, we observe the emergence of discourse-level pivot cues, which we have grouped into a turning-point set $\mathcal{T}$ and a summarization set $\mathcal{S}$. 
Semantically, tokens in $\mathcal{T}$ often signal a discourse shift that can promote continued deliberation and reflection, whereas tokens in $\mathcal{S}$ tend to indicate consolidation and closure. 
This motivates the question of whether the model, when attempting to enter (or avoid entering) a reflective regime, systematically redistributes probability mass between $\mathcal{T}$ and $\mathcal{S}$.
To quantify such an effect, we introduce the Deep-Thinking Trend (DTT) metric, measuring whether the semantic-pivot stage shifts toward $\mathcal{T}$ or toward $\mathcal{S}$. 
Concretely, we calculate the ratio of the average probabilities of two sets as:
\begin{equation}
\mathrm{DTT}^{(l)} \;=\; \frac{\overline{P_{\ell}(\mathcal{T})}}{\overline{P_{\ell}(\mathcal{S})}},
\end{equation}
where larger $\mathrm{DTT}^{(l)}$ indicates a shift toward $\mathcal{T}$, while smaller values indicate a shift toward $\mathcal{S}$.

\paragraph{Prompt-level intervention.}
As shown in Fig.~\ref{fig:piv_sem_prompt_ana}, both models exhibit a consistent trend: appending a deep-thinking cue increases DTT, shifting more probability mass toward $\mathcal{T}$, whereas appending a quick-thinking cue decreases DTT, shifting mass toward $\mathcal{S}$.
These opposite prompt effects provide initial evidence for a prompt-sensitive redistribution, indeed, a competitive trade-off between $\mathcal{T}$ and $\mathcal{S}$ within the semantic-pivot layers.


\paragraph{Activation-steering intervention.}
Activation steering provides a finer-grained test by varying the strength $\alpha$ along $\mathbf{d}_{\text{pos}}$. As shown in Fig.~\ref{fig:piv_sem_steer_ana}, steering toward the deep-thinking direction yields a monotonic increase in DTT with larger $\alpha>0$, whereas steering toward the quick-thinking direction yields a monotonic decrease in DTT as $\alpha<0$ becomes larger in magnitude. 
Together, these results provide stronger evidence that, as the model attempts to enter (or avoid entering) a reflective regime, it will induce a dynamic competition between $\mathcal{T}$ and $\mathcal{S}$ within the semantic-pivot layers, whereby probability mass is systematically shifted from one set to the other.


\subsection{Effect on Behavior-Overt Layers}

Since decoding-time sampling is typically restricted to top-ranked candidates, we follow Sec.~\ref{sec:forw_acti_ana} and track the reflection-behavior set $\mathcal{R}$, measuring how its overall probability mass $P_{\ell}(\mathcal{R})$ changes under intervention in the behavior-overt layers. In parallel, we define a non-reflection set $\mathcal{NR}$ as the remaining tokens within the Top-10 candidates and track $P_{\ell}(\mathcal{NR})$. Notably, $\mathcal{NR}$ often contains tokens such as ``So'' or ``I'', which are characteristic of a regime that avoids entering reflective behavior.
The corresponding defined token sets can be found in App.~\ref{app_pre_difine_tokens}.

\begin{figure*}[ht]
  \centering
    \includegraphics[width=0.95\linewidth]{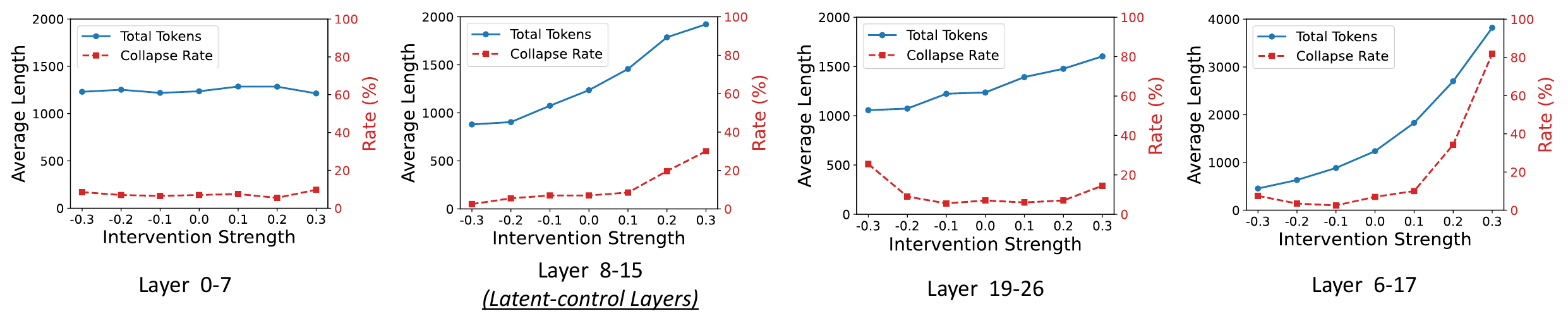}
  \caption{Effect on overall thinking length under DeepSeek-R1$_{7B}$ with activation-steering interventions, and the analysis on Qwen3-Think$_{4B}$ can be found in Fig.~\ref{fig:overall_length_qw3}. We report the average response length as well as the collapse rate caused by degraded language ability.}
  \label{fig:overall_length_dsqw}
  \vspace{-0.4cm}
\end{figure*}

\paragraph{Prompt-level intervention.}
As shown in Fig.~\ref{fig:behvave_prompt_ana}, both models exhibit a consistent trend: appending a deep-thinking cue increases $P_{\ell}(\mathcal{R})$ while decreasing $P_{\ell}(\mathcal{NR})$, indicating a higher likelihood of entering a reflective regime. Conversely, appending a quick-thinking cue decreases $P_{\ell}(\mathcal{R})$ and increases $P_{\ell}(\mathcal{NR})$, suggesting a reduced tendency to enter reflection. These results indicate that shifts induced in earlier stages ultimately manifest as measurable changes in reflection-related behavior in the behavior-overt stage.

\paragraph{Activation-steering intervention.}
Fig.~\ref{fig:behvave_steer_ana} shows the same qualitative pattern under activation steering. Steering toward the deep-thinking direction increases $P_{\ell}(\mathcal{R})$ and decreases $P_{\ell}(\mathcal{NR})$, with stronger effects at larger intervention magnitudes; steering toward the quick-thinking direction yields the opposite trend. Together, these results further support that intervention effects propagate to the behavior-overt layers and directly modulate the likelihood of entering an explicit reflective regime.

Overall, this section traces how each stage responds to targeted interventions, thereby validating the functional roles of different layer groups and supporting a coherent causal chain across depth. 
Concretely, prompt-level semantics first modulate the projection of activations onto the latent-control residual direction. 
This perturbation then induces competition between turning-point and summarization cues in the semantic-pivot layers, and ultimately changes the sampling likelihood of reflection tokens in the behavior-overt layers.

\section{Discussion}

\begin{figure}[t]
  \centering
  \begin{subfigure}[t]{0.24\textwidth}
    \centering
    \includegraphics[width=\linewidth]{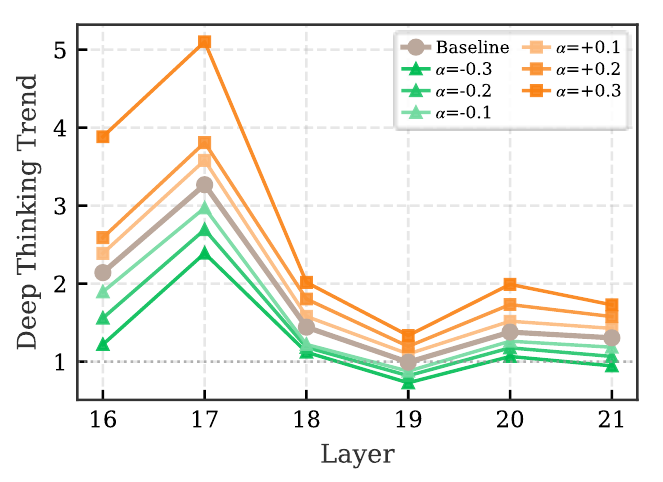}
    \caption{Analysis on ``Hmm'' token.}
      \label{dsqw_hmm_dtt}
  \end{subfigure}\hfill
  \begin{subfigure}[t]{0.24\textwidth}
    \centering
    \includegraphics[width=\linewidth]{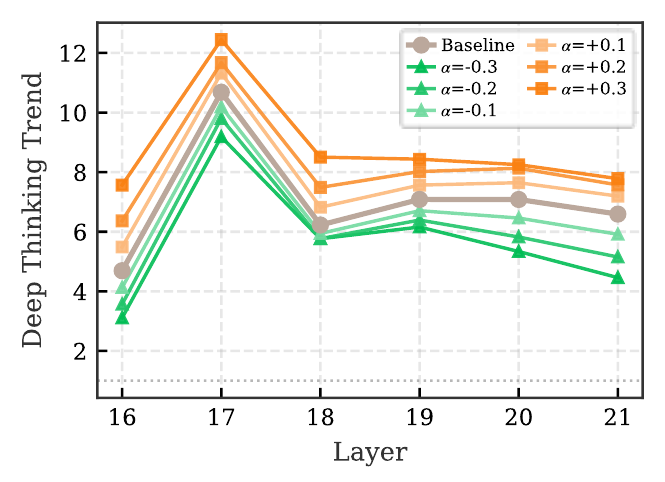}
    \caption{Analysis on medical data.}
    \label{dsqw_med_dtt}
  \end{subfigure}
  \caption{Effect on semantic-pivot layers under DeepSeek-R1$_{7B}$ with activation-steering interventions.}
  \vspace{-0.6cm}
\end{figure}

\subsection{How do activation-steering interventions in the latent-control layers affect the thinking length?}\label{sec_ana_act_steer}

Our study further analyzes the stability and specificity of activation steering on latent-control layers. 
First, we observe that overly large steering magnitudes will cause severe language degradation, which motivates restricting $\alpha$ to a stable range in all experiments. 
Second, we compare steering different layer groups and their effects on thinking length. 
As shown in Fig.~\ref{fig:overall_length_dsqw} and \ref{fig:overall_length_qw3} (in our appendix), intervening on latent-control layers always yields the strong and reliable linear control over thinking length, while steering comparable other layers groups is less effective. 
Moreover, extending the intervention beyond the latent-control layers by including additional neighboring layers will increase the risk of language collapse. 
Together, these results suggest that the latent-control layers play a critical role in regulating the thinking budget.

\subsection{Does the stage-wise pattern persist when anchoring on other reflection markers?}
Our main analyses align the reflection onset by anchoring on ``Wait''. 
To test whether the stage-wise pattern depends on this specific marker, we additionally anchor on ``Hmm'' and repeat the activation-steering experiments on DeepSeek-R1$_{7B}$ using GSM8K, resulting in 371 anchored samples. 
As shown in Figs.~\ref{dsqw_hmm_dtt} and \ref{dsqw_hmm_behave} (in our appendix), steering toward deep thinking ($\alpha>0$) increases DTT and the sampling likelihood of reflection-behavior tokens, while steering toward quick thinking ($\alpha<0$) yields the opposite trend. 
These results closely match the ``Wait''-anchored findings, indicating that the mechanism is not tied to a single marker.

\subsection{Does the stage-wise pattern generalize to other domains?}
To assess domain generalization beyond GSM8K, we conduct the same activation-steering analysis on MedMCQA~\cite{pal2022medmcqa} with DeepSeek-R1$_{7B}$, resulting in 731 samples anchored on ``Wait'' token onset. 
As shown in Figs.~\ref{dsqw_med_dtt} and \ref{dsqw_med_behave} (in our appendix), we observe the same monotonic relationship between steering strength and both DTT and reflection-token likelihood: $\alpha>0$ increases them, whereas $\alpha<0$ decreases them. This suggests that the identified stage-wise pattern generalizes to various domains.

\section{Conclusion and Future Applications}

In this study, our correlation and intervention analyses jointly reveal a structured, stage-wise progression across depth as reflective behavior emerges. 
These findings suggest that reflection is supported by salient internal signals that can be traced and causally influenced, offering a principled basis for predicting, controlling, and improving reflective behaviors in R1-style LLMs. 
Building on this view, future work may (i) regulate reflection via quantitative interventions in latent-control layers, (ii) forecast reflection by modeling activation-shift features and their trajectories, and (iii) enhance reflection by injecting supervision of varying granularity into intermediate layers.



\section{Impact Statements}

This work advances the mechanistic understanding of how reflection behavior is governed inside large language models. 
Combining t-SNE visualization, logit-lens decoding of activation differences, and targeted causal interventions, we identify depth intervals with distinct functional roles that mediate a progression from prompt-level control to discourse-level semantics and, ultimately, overt behavior. 
We further introduce simple metrics and reproducible protocols for tracking and manipulating these internal signals across models, supporting future research on interpretability and controlled generation. 
This study is primarily analytical and does not propose deployment-facing systems. 
Accordingly, we do not anticipate significant direct negative societal impacts.

\bibliography{custom}
\bibliographystyle{icml2026}

\newpage
\appendix
\onecolumn




\section{Additional Visualizations}
In our appendix, we further provide supplementary visualizations that support the layer-wise analysis in the main paper.
We include three complementary views for each model: (i) T-SNE visualization of per-layer hidden activations, (ii) decoding results of per-layer activation differences, and (iii) decoding results of per-layer forward activations.


\subsection{T-SNE Visualization}\label{app_visual_tsne}
\begin{figure}[t]
  \centering
  \makebox[\linewidth][c]{\includegraphics[width=\linewidth,height=0.78\textheight,keepaspectratio]{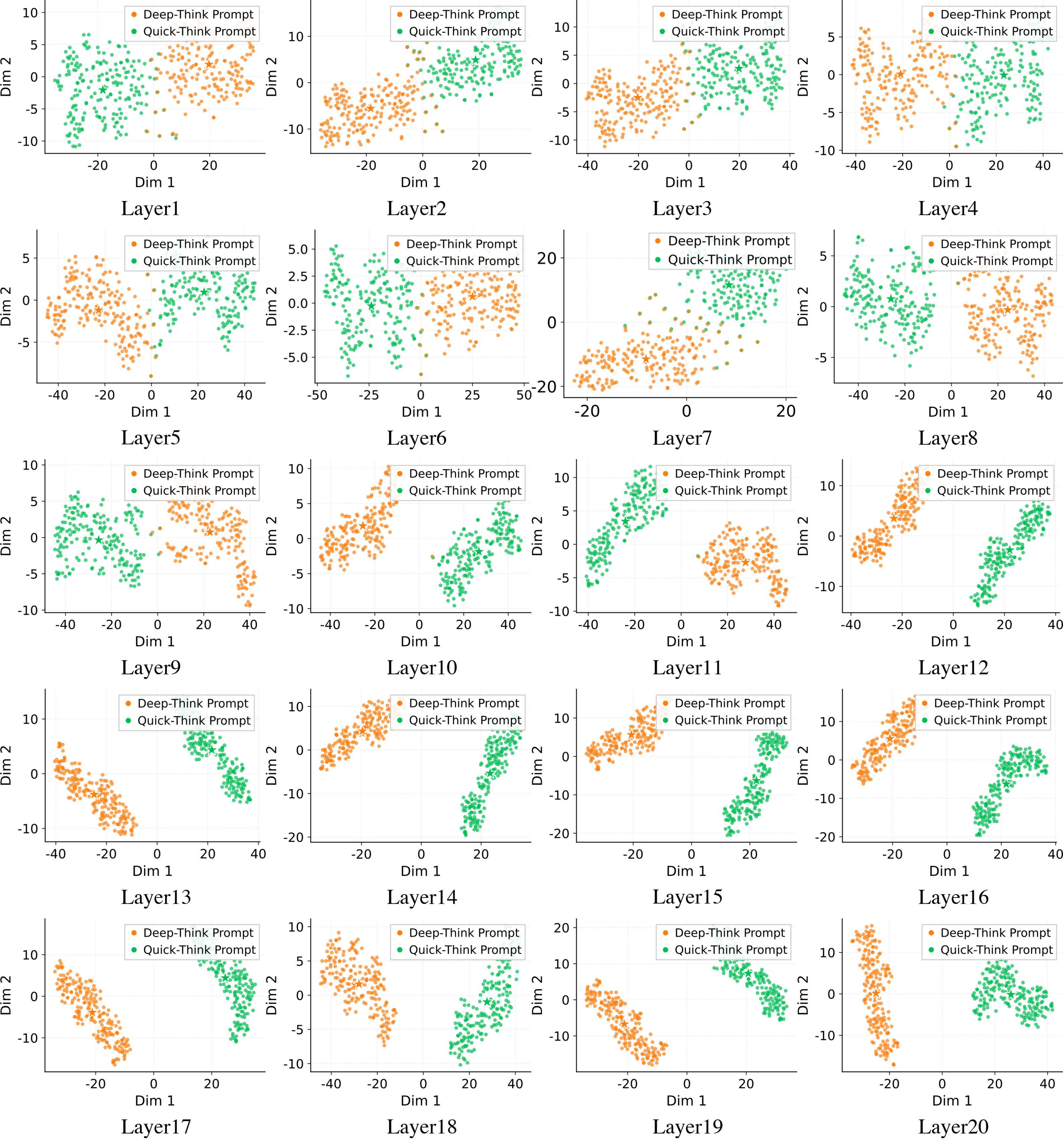}}
  \caption{T-SNE visualization of layer-wise activations for DeepSeek-R1$_{\text{7B}}$ under Quick-Think vs.\ Deep-Think prompts (\emph{Part} 1).}
  \label{fig:tsne-ds-part1}
\end{figure}

\begin{figure}[t]
  \centering
  \makebox[\linewidth][c]{\includegraphics[width=\linewidth,height=0.78\textheight,keepaspectratio]{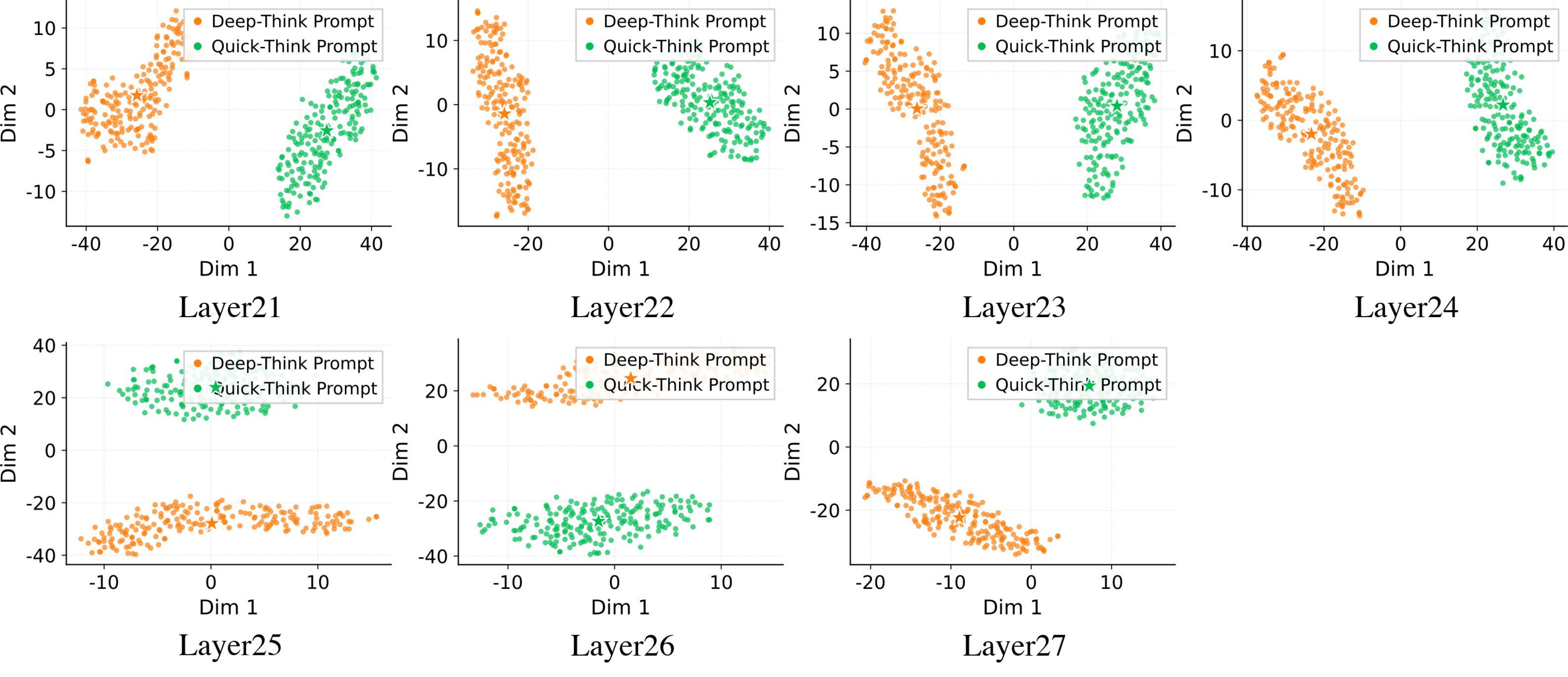}}
  \caption{T-SNE visualization of layer-wise activations for DeepSeek-R1$_{\text{7B}}$ under Quick-Think vs.\ Deep-Think prompts (\emph{Part} 2).}
  \label{fig:tsne-ds-part2}
\end{figure}

\begin{figure}[t]
  \centering
  \makebox[\linewidth][c]{\includegraphics[width=\linewidth,height=0.78\textheight,keepaspectratio]{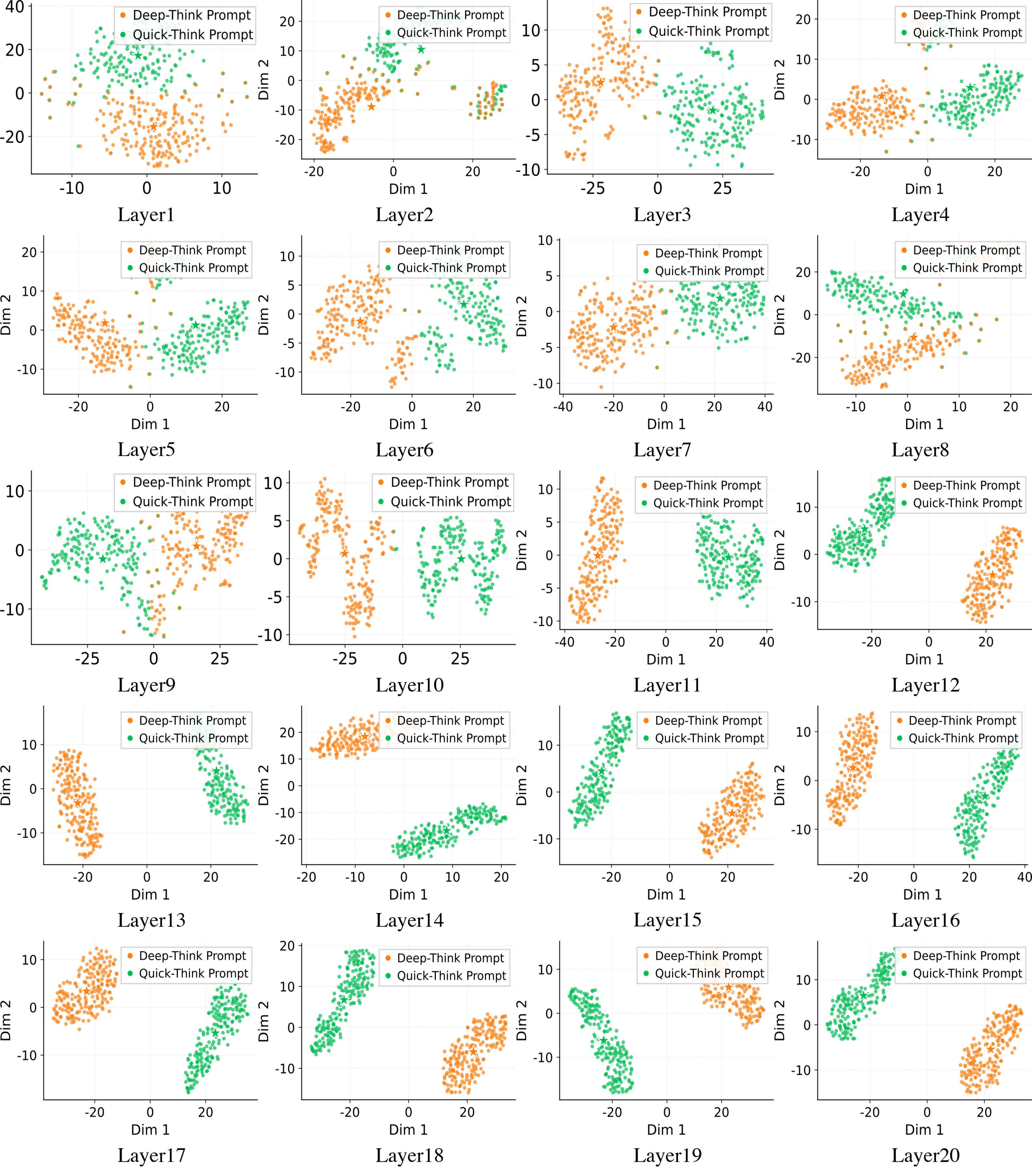}}
  \caption{T-SNE visualization of layer-wise activations for Qwen3-Think$_{\text{4B}}$ under Quick-Think vs.\ Deep-Think prompts (\emph{Part} 1).}
  \label{fig:tsne-qw-part1}
\end{figure}

\begin{figure}[t]
  \centering
  \makebox[\linewidth][c]{\includegraphics[width=\linewidth,height=0.78\textheight,keepaspectratio]{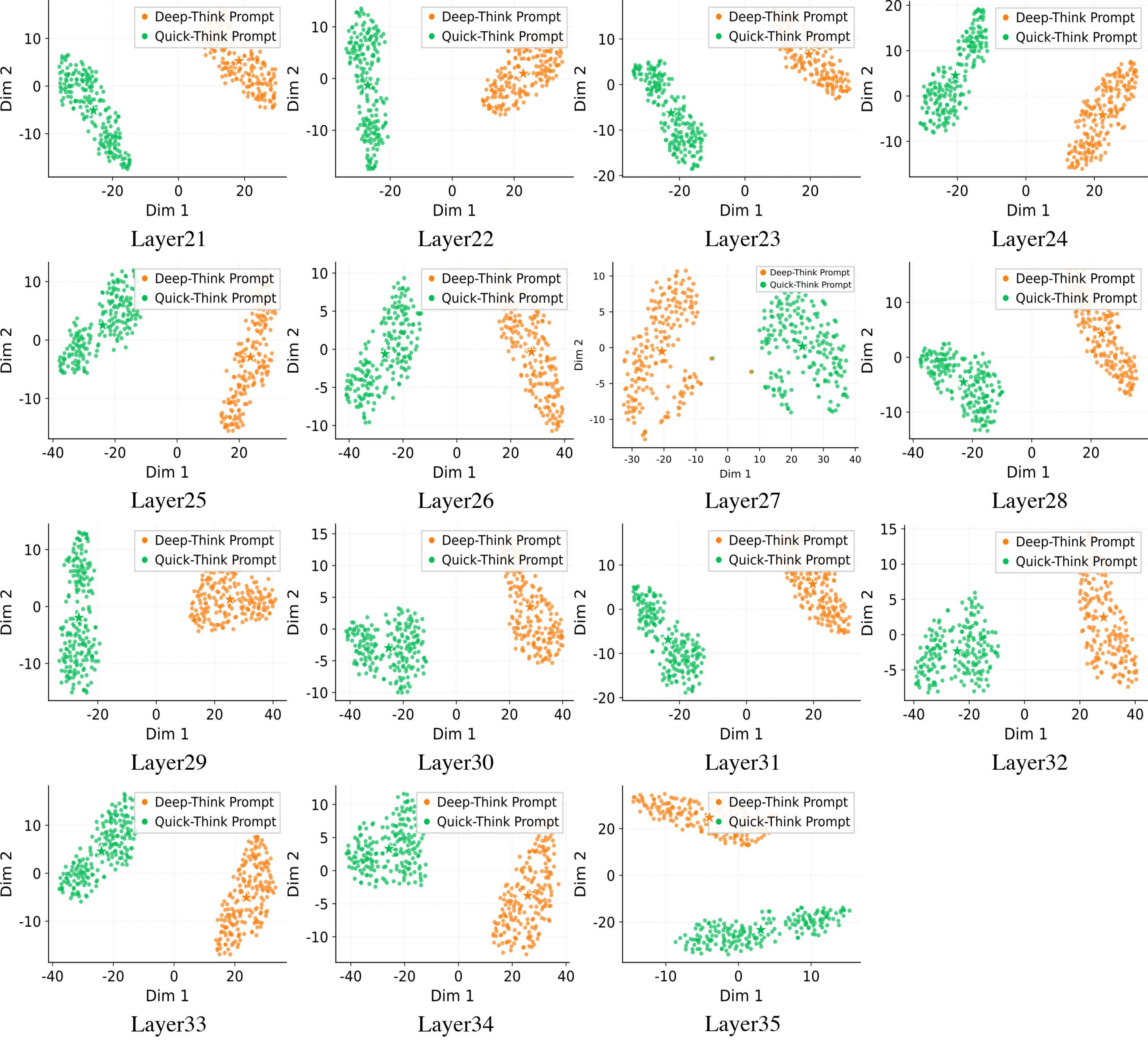}}
  \caption{T-SNE visualization of layer-wise activations for Qwen3-Think$_{\text{4B}}$ under Quick-Think vs.\ Deep-Think prompts (\emph{Part} 2).}
  \label{fig:tsne-qw-part2}
\end{figure}

Figs.~\ref{fig:tsne-ds-part1}-\ref{fig:tsne-qw-part2} report results for DeepSeek-R1$_{\text{7B}}$ and Qwen3-Think$_{\text{4B}}$.
We apply T-SNE to visualize hidden activations at each layer under Quick-Think and Deep-Think prompts, and clearer clustering (or larger separation) indicates stronger prompt-induced representational divergence.

\subsection{Decoding Results of Activation Differences}\label{app_visual_diff_decode}
\begin{figure}[t]
  \centering
  \makebox[\linewidth][c]{\includegraphics[width=\linewidth,height=0.78\textheight,keepaspectratio]{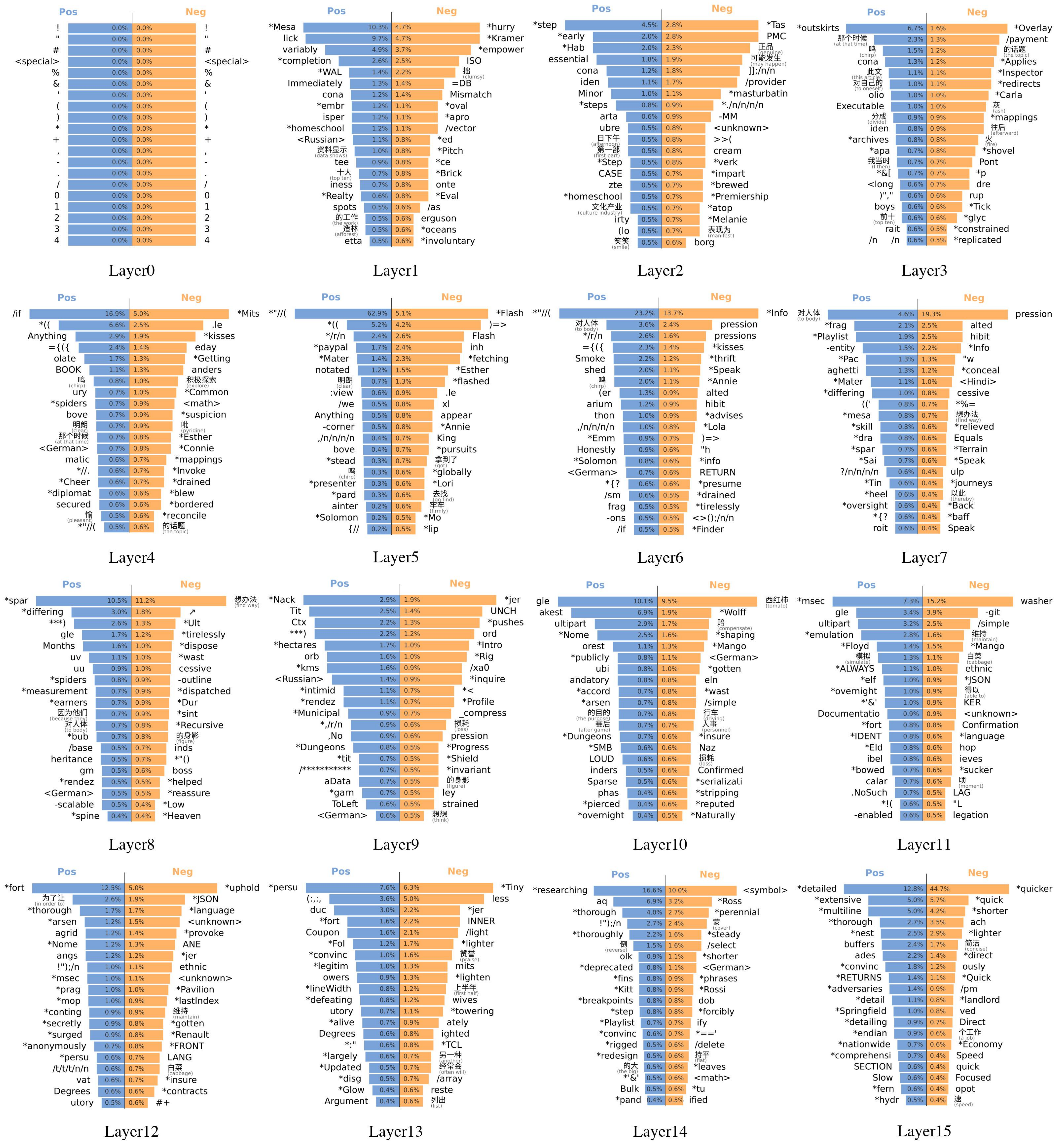}}
  \caption{Decoding results of activation differences for DeepSeek-R1$_{\text{7B}}$ (\emph{Part} 1). The asterisk (*) next to a token denotes a placeholder for a whitespace character.}
  \label{fig:butterfly-ds-part1}
\end{figure}

\begin{figure}[t]
  \centering
  \makebox[\linewidth][c]{\includegraphics[width=\linewidth,height=0.78\textheight,keepaspectratio]{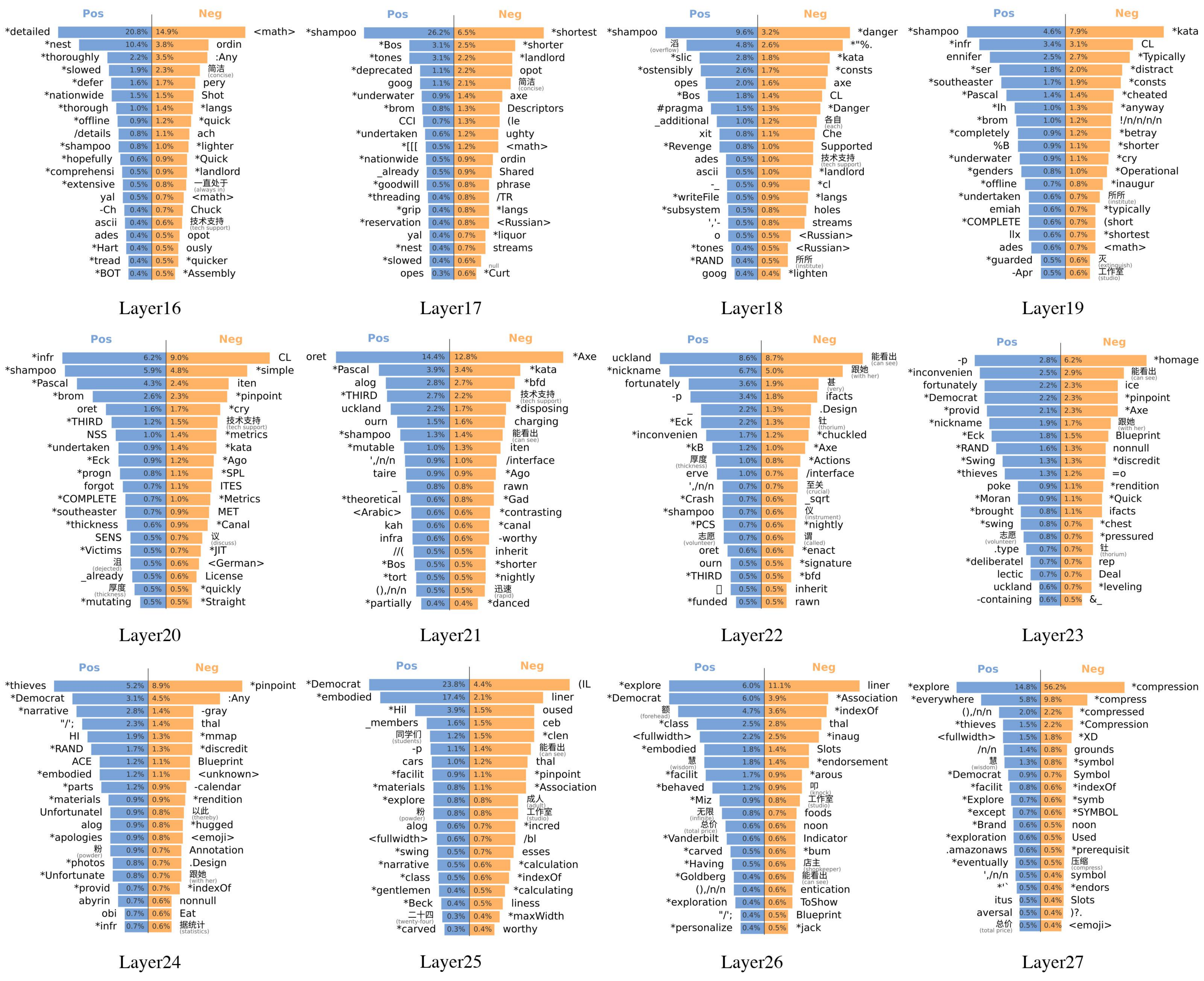}}
  \caption{Decoding results of activation differences for DeepSeek-R1$_{\text{7B}}$ (\emph{Part} 2). The asterisk (*) next to a token denotes a placeholder for a whitespace character.}
  \label{fig:butterfly-ds-part2}
\end{figure}

\begin{figure}[t]
  \centering
  \makebox[\linewidth][c]{\includegraphics[width=\linewidth,height=0.78\textheight,keepaspectratio]{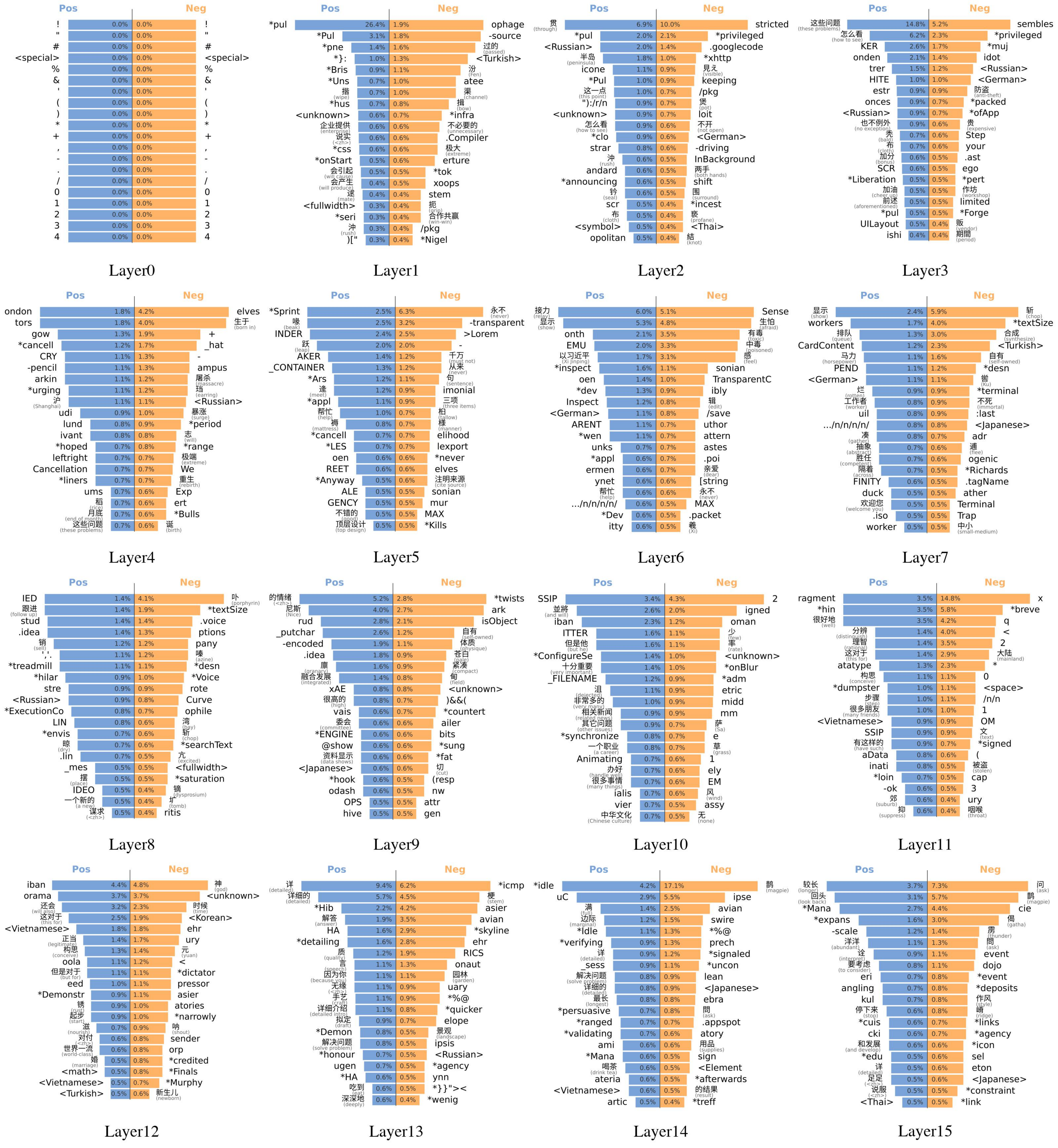}}
  \caption{Decoding results of activation differences for Qwen3-Think$_{\text{4B}}$ (\emph{Part} 1). The asterisk (*) next to a token denotes a placeholder for a whitespace character.}
  \label{fig:butterfly-qw-part1}
\end{figure}

\begin{figure}[t]
  \centering
  \makebox[\linewidth][c]{\includegraphics[width=\linewidth,height=0.78\textheight,keepaspectratio]{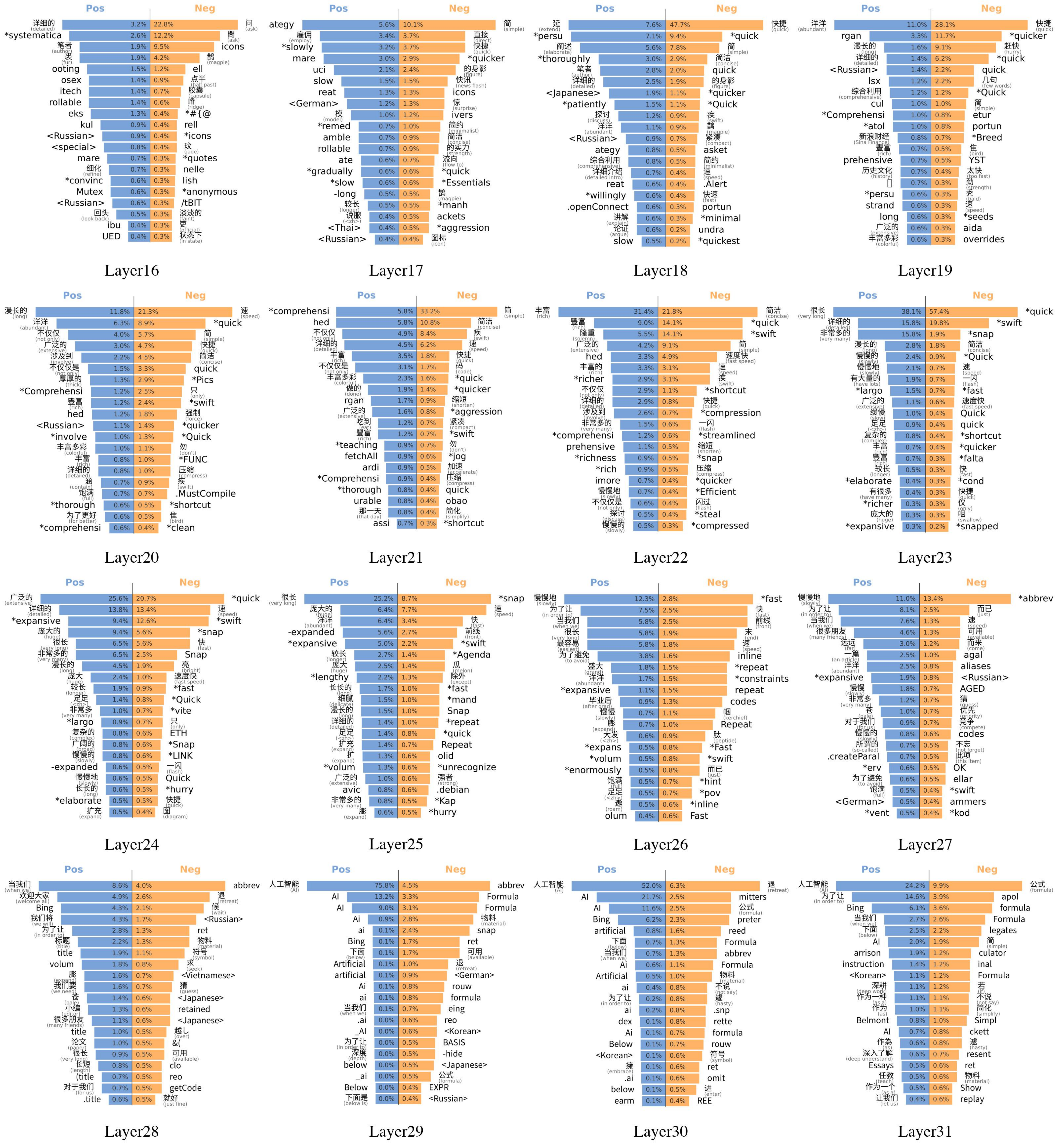}}
  \caption{Decoding results of activation differences for Qwen3-Think$_{\text{4B}}$ (\emph{Part} 2). The asterisk (*) next to a token denotes a placeholder for a whitespace character.}
  \label{fig:butterfly-qw-part2}
\end{figure}

\begin{figure}[t]
  \centering
  \makebox[\linewidth][c]{\includegraphics[width=\linewidth,height=0.78\textheight,keepaspectratio]{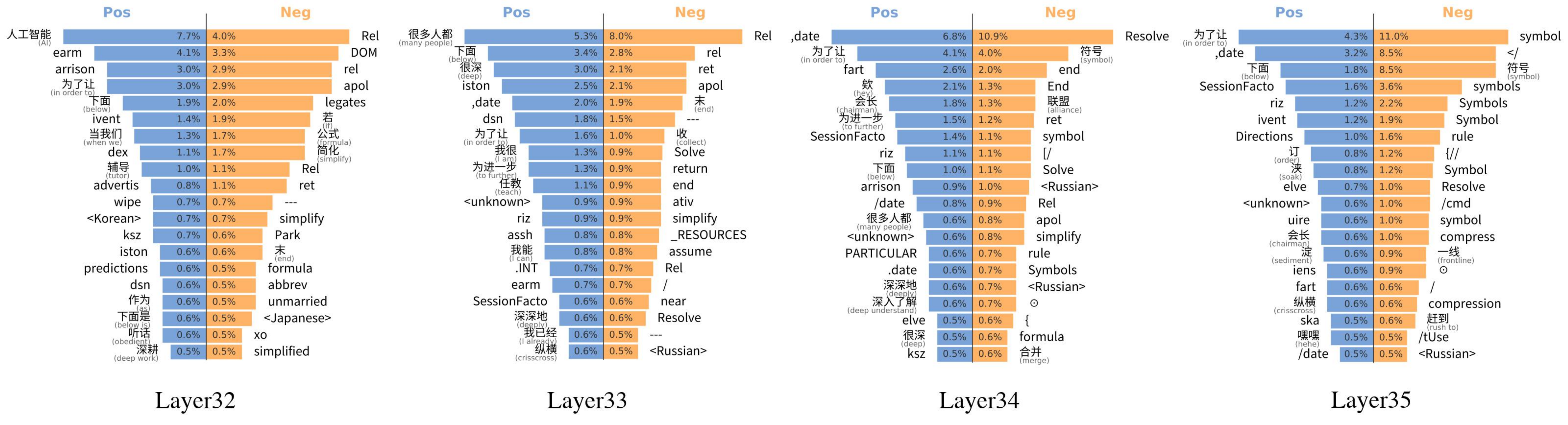}}
  \caption{Decoding results of activation differences for Qwen3-Think$_{\text{4B}}$ (\emph{Part} 3). The asterisk (*) next to a token denotes a placeholder for a whitespace character.}
  \label{fig:butterfly-qw-part3}
\end{figure}

Figs.~\ref{fig:butterfly-ds-part1}-\ref{fig:butterfly-qw-part3} report decoding results of layer-wise activation differences  for DeepSeek-R1$_{\text{7B}}$ and Qwen3-Think$_{\text{4B}}$.
These plots reveal how thinking-budget semantics evolve across depth.

\subsection{Decoding Results of Forward Activation}\label{app_visual_forward_decode}
\begin{figure}[t]
  \centering
  \makebox[\linewidth][c]{\includegraphics[width=\linewidth,height=0.78\textheight,keepaspectratio]{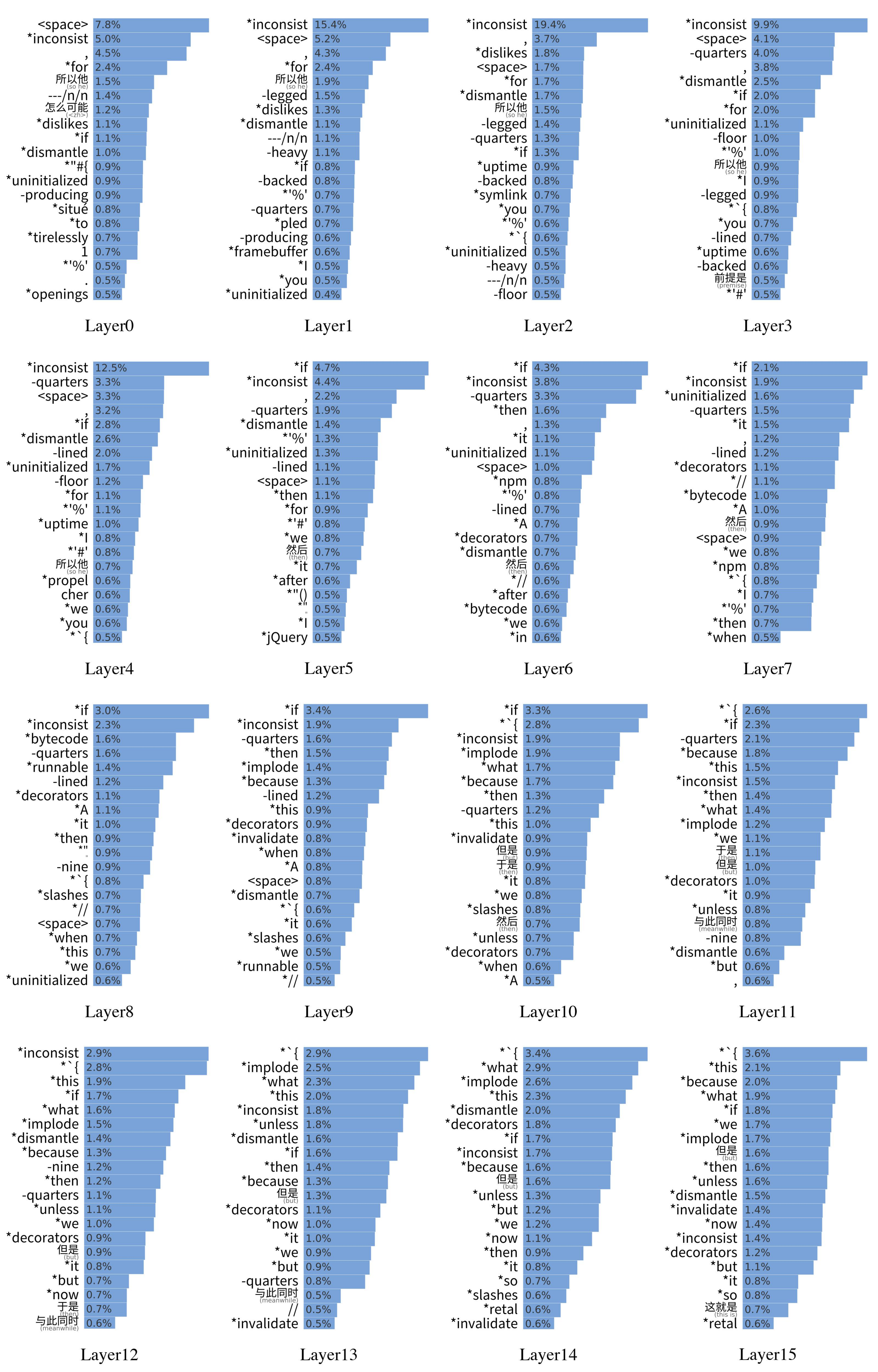}}
  \caption{Decoding results of forward activations for DeepSeek-R1$_{\text{7B}}$ (\emph{Part} 1). The asterisk (*) next to a token denotes a placeholder for a whitespace character.}
  \label{fig:top50-ds-part1}
\end{figure}

\begin{figure}[t]
  \centering
  \makebox[\linewidth][c]{\includegraphics[width=0.8\linewidth,height=0.78\textheight,keepaspectratio]{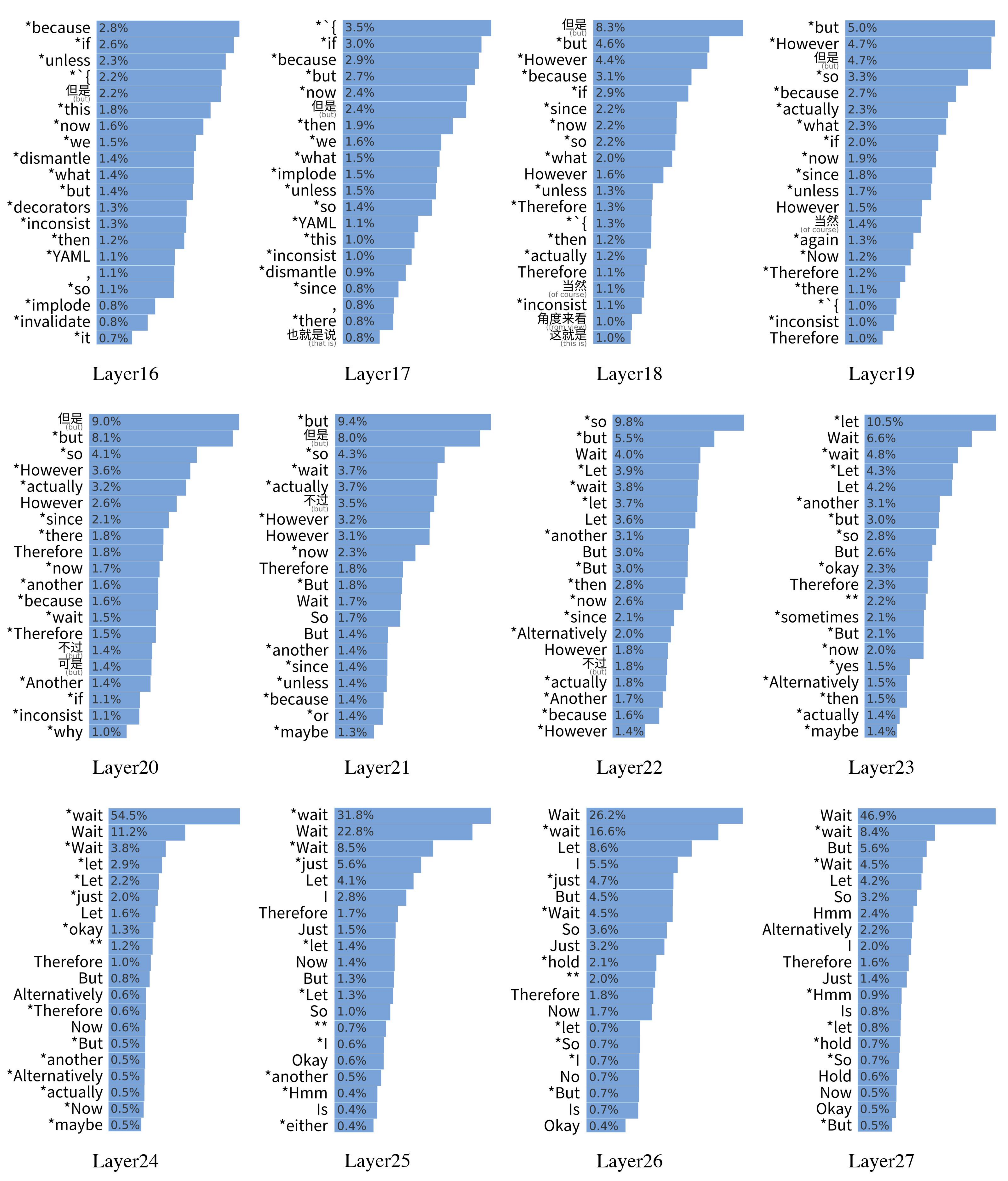}}
  \caption{Decoding results of forward activations for DeepSeek-R1$_{\text{7B}}$ (\emph{Part} 2). The asterisk (*) next to a token denotes a placeholder for a whitespace character.}
  \label{fig:top50-ds-part2}
\end{figure}

\begin{figure}[t]
  \centering
  \makebox[\linewidth][c]{\includegraphics[width=\linewidth,height=0.78\textheight,keepaspectratio]{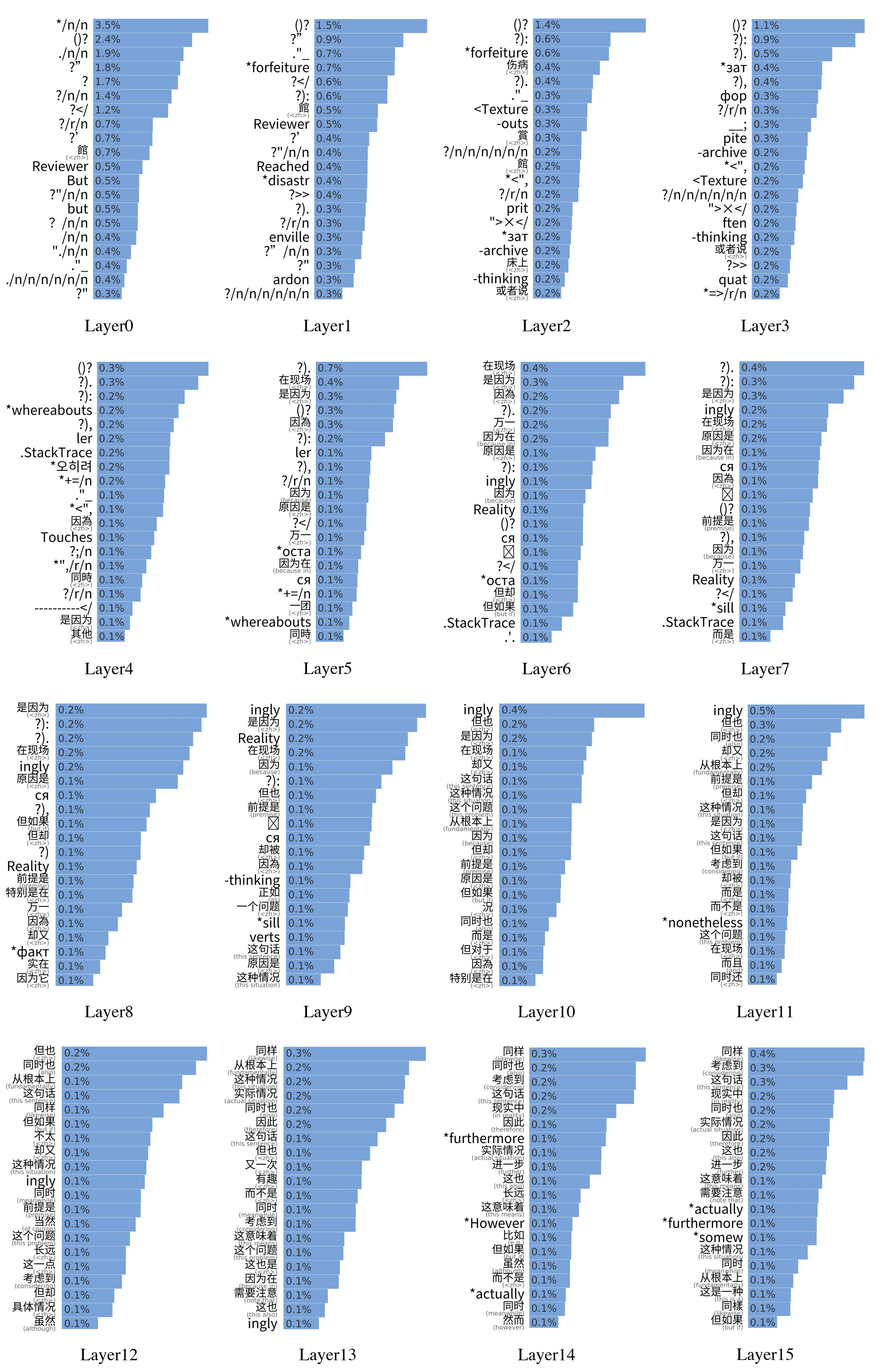}}
  \caption{Decoding results of forward activations for Qwen3-Think$_{\text{4B}}$ (\emph{Part} 1). The asterisk (*) next to a token denotes a placeholder for a whitespace character.}
  \label{fig:top50-qw-part1}
\end{figure}

\begin{figure}[t]
  \centering
  \makebox[\linewidth][c]{\includegraphics[width=\linewidth,height=0.78\textheight,keepaspectratio]{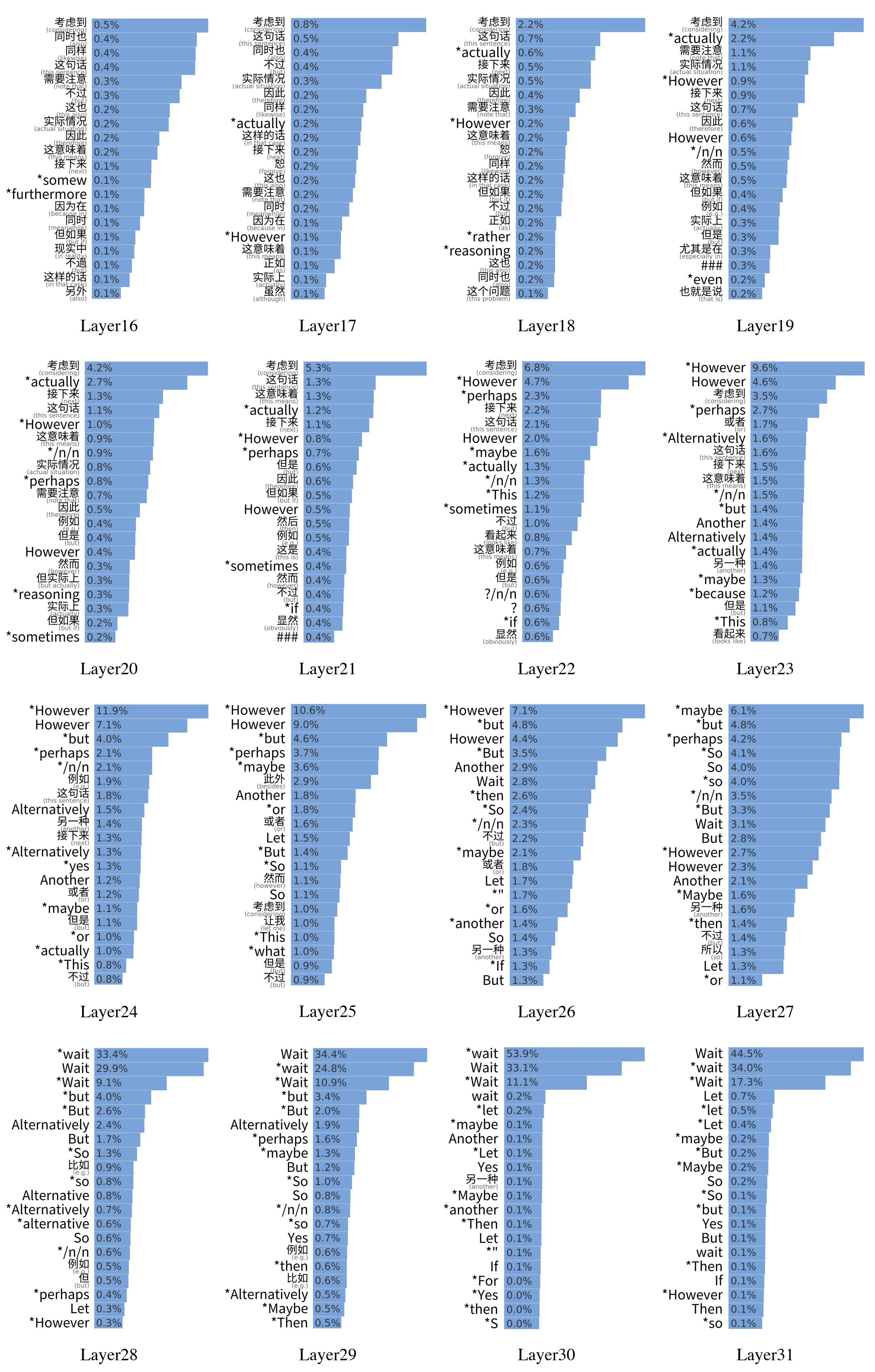}}
  \caption{Decoding results of forward activations for Qwen3-Think$_{\text{4B}}$ (\emph{Part} 2). The asterisk (*) next to a token denotes a placeholder for a whitespace character.}
  \label{fig:top50-qw-part2}
\end{figure}

\begin{figure}[t]
  \centering
  \makebox[\linewidth][c]{\includegraphics[width=0.8\linewidth,height=0.6\textheight,keepaspectratio]{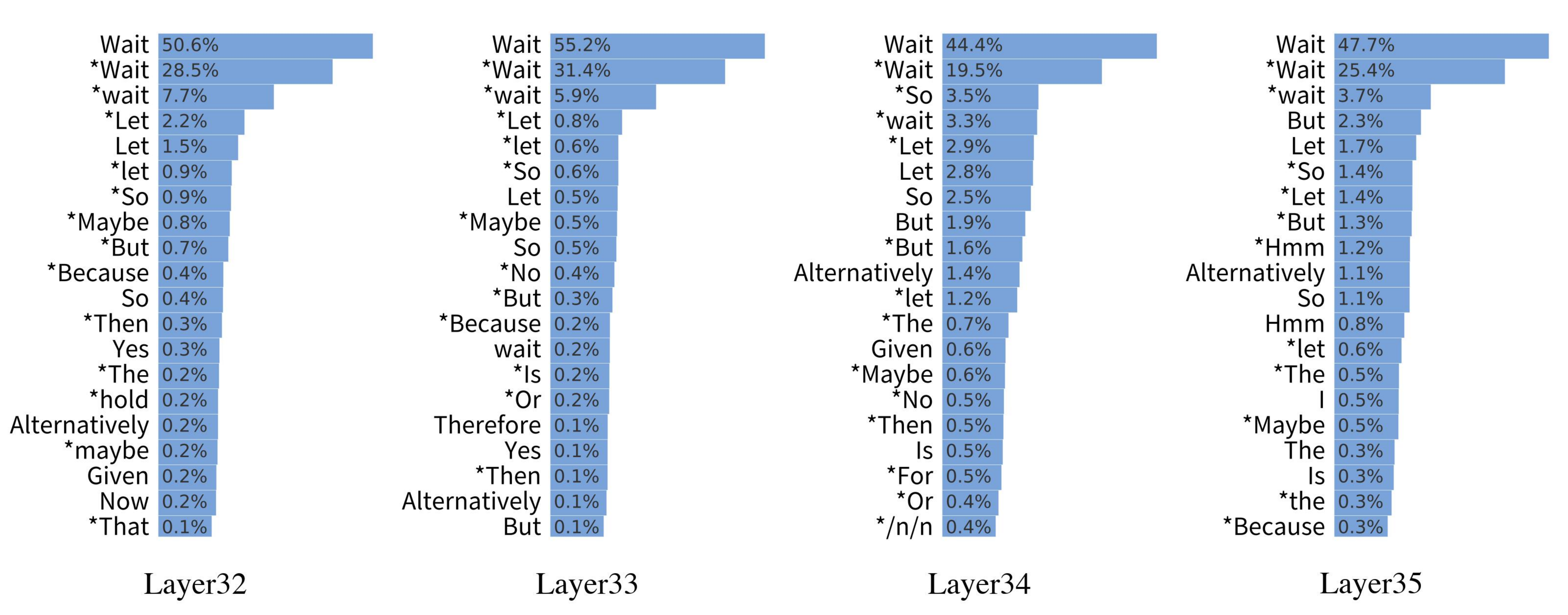}}
  \caption{Decoding results of forward activations for Qwen3-Think$_{\text{4B}}$ (\emph{Part} 3). The asterisk (*) next to a token denotes a placeholder for a whitespace character.}
  \label{fig:top50-qw-part3}
\end{figure}

\noindent
Figs.~\ref{fig:top50-ds-part1}-\ref{fig:top50-qw-part3} report decoding results of layer-wise forward activations for DeepSeek-R1$_{\text{7B}}$ and Qwen3-Think$_{\text{4B}}$.
This provides a depth-wise view of how reflection-related signals amplify (or attenuate) as information propagates forward.


\section{Pre-defined Token Set}\label{app_pre_difine_tokens}
\noindent
We provide the pre-defined token sets used throughout the main text, including turning-point tokens ($\mathcal{T}$), summarization tokens ($\mathcal{S}$), reflection tokens ($\mathcal{R}$), and non-reflection tokens ($\mathcal{NR}$).

For DeepSeek-R1$_{\text{7B}}$, the corresponding set includes:
\begin{itemize}[leftmargin=*,noitemsep,topsep=2pt]
  \item \textbf{Turning-point tokens ($\mathcal{T}$):}
  \texttt{` but', ` But', `But', `However', ` However', ` however', ` Let', ` let', `Let', ` perhaps', ` actually', `Wait', ` wait', \begin{CJK*}{UTF8}{gbsn}`可是' (but), `但在' (but in), `但是' (but), `不过' (however)\end{CJK*}}.
  \item \textbf{Summarization tokens ($\mathcal{S}$):}
  \texttt{` So', `So', `so', `Therefore', \begin{CJK*}{UTF8}{gbsn}`所以' (so), `所以说' (so), `所以在' (so in), `这就是' (this is)\end{CJK*}}.
  \item \textbf{Reflection tokens ($\mathcal{R}$):}
  \texttt{`Wait', `wait', `But', ` Wait', `Let', `Hmm'}.
  \item \textbf{Non-Reflection tokens ($\mathcal{NR}$):}
  \texttt{`So', `Alternatively', `I', `Therefore'}.
\end{itemize}


And for Qwen3-Think$_{\text{4B}}$, the corresponding set includes:
\begin{itemize}[leftmargin=*,noitemsep,topsep=2pt]
  \item \textbf{Turning-point tokens ($\mathcal{T}$):}
  \texttt{` However', `However', ` But', `Wait', \begin{CJK*}{UTF8}{gbsn}`不过' (however)\end{CJK*}, `Let', ` another', `But', ` perhaps', \begin{CJK*}{UTF8}{gbsn}`但是' (but)\end{CJK*}, ` Let', \begin{CJK*}{UTF8}{gbsn}`让我' (let me)\end{CJK*}, ` wait', ` Wait', ` Hmm', ` actually', ` let', ` however', ` but', \begin{CJK*}{UTF8}{gbsn}`然而' (however)\end{CJK*}}.
  \item \textbf{Summarization tokens ($\mathcal{S}$):}
  \texttt{`So', ` So', ` so', `Therefore', ` Therefore', \begin{CJK*}{UTF8}{gbsn}`因此' (therefore), `所以' (so), `这意味着' (this means)\end{CJK*}}.
  \item \textbf{Reflection tokens ($\mathcal{R}$):}
  \texttt{`Wait', `wait', `But', ` Wait', `Let', `Hmm', ` Let', ` But'}.
  \item \textbf{Non-Reflection tokens ($\mathcal{NR}$):}
  \texttt{`Alternatively', ` So'}.
\end{itemize}


\begin{figure*}[t]
  \centering
    \includegraphics[width=0.95\linewidth]{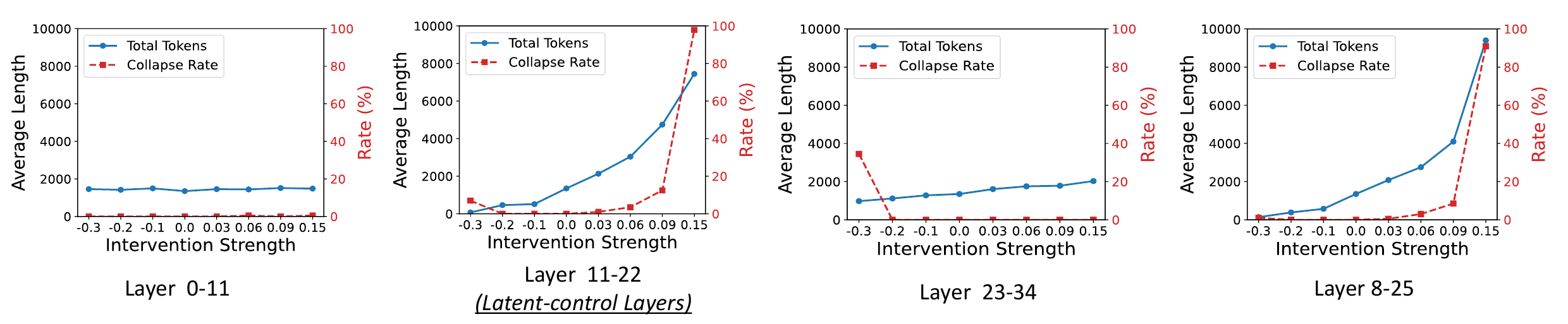}
  \caption{Effect on overall thinking length under Qwen3-Think$_{4B}$ with activation-steering interventions.}
  \label{fig:overall_length_qw3}
\end{figure*}

\begin{figure}[t]
  \centering
  \begin{subfigure}[t]{0.47\textwidth}
    \centering
    \includegraphics[width=\linewidth]{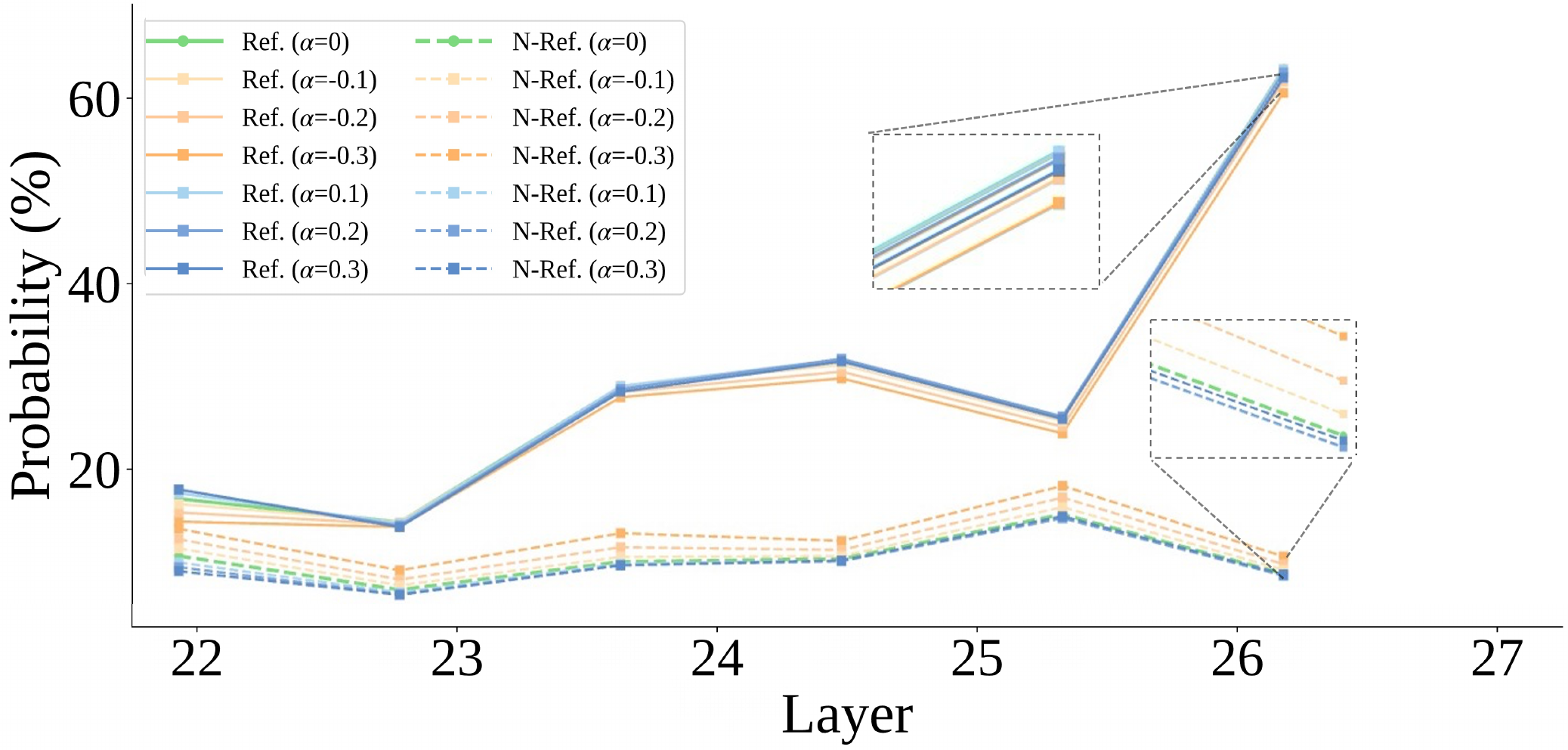}
    \caption{Analysis on ``Hmm'' token.}
    \label{dsqw_hmm_behave}
  \end{subfigure}\hfill
  \begin{subfigure}[t]{0.43\textwidth}
    \centering
    \includegraphics[width=\linewidth]{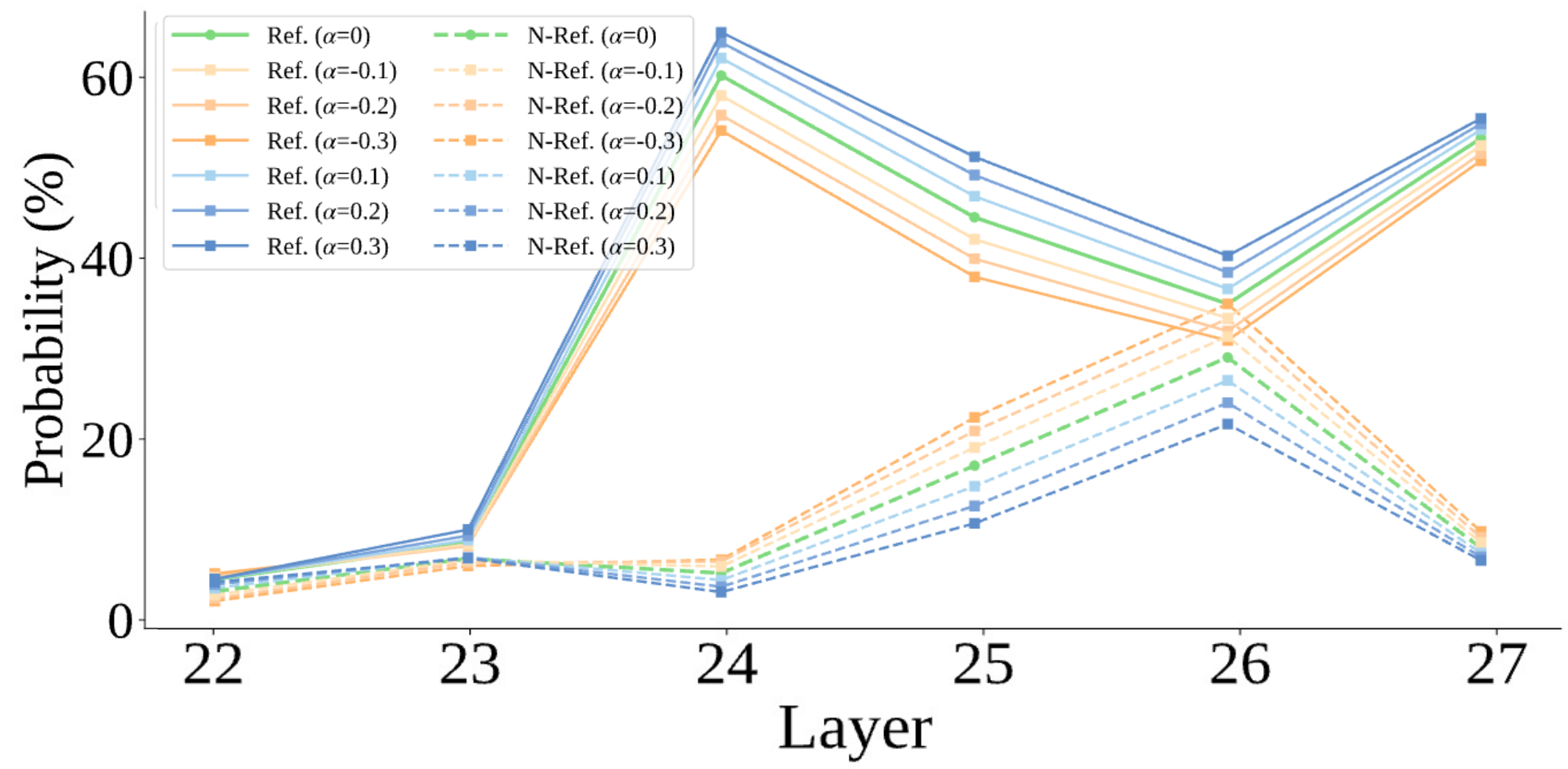}
    \caption{Analysis on medical data.}
    \label{dsqw_med_behave}
  \end{subfigure}
  \caption{Effect on behavior-overt layers under DeepSeek-R1$_{7B}$ with activation-steering interventions. Ref. and N-Ref. represent $\mathcal{R}$ and $\mathcal{NR}$ sets.}
\end{figure}

\clearpage
\onecolumn

\end{document}